\documentclass[review,3p]{elsarticle}

\usepackage{amsmath}
\usepackage{multirow}
\usepackage{textcomp}
\usepackage{enumerate}
\usepackage{amssymb}
\usepackage[usenames, dvipsnames]{color}
\usepackage{lscape}
\usepackage{rotating}
\usepackage{tikz}
\usepackage{hyperref}
\usepackage[caption=false,font=footnotesize]{subfig}
\usepackage{algorithm}
\usepackage[noend]{algpseudocode}
\usepackage{algorithmicx}
\usepackage{rotating}
\usepackage{booktabs}
\usepackage{tabularx}
\usepackage[acronym, nonumberlist]{glossaries}

\makenoidxglossaries

\newacronym{ml}{ML}{Machine Learning}
\newacronym{svm}{SVM}{Support Vector Machine}
\newacronym{hp}{HP}{Hyperparameter}
\newacronym{mtl}{MtL}{Meta-learning}
\newacronym{gs}{GS}{Grid Search}
\newacronym{rs}{RS}{Random Search}
\newacronym{smbo}{SMBO}{Sequential Model-based Optimization}
\newacronym{openml}{OpenML}{Open Machine Learning}
\newacronym{cv}{CV}{Cross-validation}
\newacronym{nmse}{NMSE}{Normalized Mean Squared Error}
\newacronym{knn}{kNN}{k-Nearest Neighbors}
\newacronym{acc}{Acc}{Accuracy}
\newacronym{at}{AT}{Active Testing}
\newacronym{pmcc}{PMCC}{Pearson Product-Moment Correlation Coefficient}
\newacronym{nae}{NAE}{Normalized Absolute Error}
\newacronym{mad}{MAD}{Mean Absolute Deviation}
\newacronym{pso}{PSO}{Particle Swarm Optimization}
\newacronym{ts}{TS}{Tabu Search}
\newacronym{ga}{GA}{Genetic Algorithm}
\newacronym{auc}{AUC}{Area Under the ROC curve}
\newacronym{bac}{BAC}{Balanced per class Accuracy}
\newacronym{cart}{CART}{Classification and Regression Tree}
\newacronym{dt}{DT}{Decision Tree}
\newacronym{dtia}{DTIA}{Decision Tree Induction Algorithm}
\newacronym{nb}{NB}{Na{\"i}ve-Bayes}
\newacronym{rf}{RF}{Random Forest}
\newacronym{lr}{LR}{Logistic Regression}
\newacronym{gp}{GP}{Gaussian Process}
\newacronym{sfs}{SFS}{Sequential Forward Selection}
\newacronym{cd}{CD}{Critical Difference}
\newacronym{eda}{EDA}{Estimation of Distribution Algorithm}
\newacronym{uci}{UCI}{University of California Irvine}
\newacronym{pca}{PCA}{Principal Component Analysis}
\newacronym{fp}{FP}{False Positive}
\newacronym{fn}{FN}{False Negative}
\newacronym{cfs}{CFS}{Correlation-based feature selection}
\newacronym{rbf}{RBF}{Radial Basis Function}
\newacronym{smote}{SMOTE}{Synthetic Minority Over-sampling Technique}
\newacronym{ds}{DS}{Decision Stump}
\newacronym{svd}{SVD}{Singular-value Decomposition}
\newacronym{taf}{TAF}{Transfer Acquisition Function}
\newacronym{irace}{Irace}{Iterated F-race}
\newacronym{roar}{ROAR}{Random Online Adaptive Racing}
\newacronym{pils}{ParamILS}{Iterated Local Search in Parameter Configuration Space}
\newacronym{scrr}{SpCorr}{Spearman Correlation}
\newacronym{mlp}{MLP}{Multilayer Perceptron}
\newacronym{cash}{CASH}{Combined Algorithm Selection and Hyperparameter Optimization}

\newacronym{rl}{RL}{Relative Landmarking}
\newacronym{sm}{SM}{Simple}
\newacronym{st}{ST}{Statistical}
\newacronym{in}{IN}{Information-theoretic}
\newacronym{mb}{MB}{Model-based}
\newacronym{lm}{LM}{Landmarking}
\newacronym{dc}{DC}{Data Complexity}
\newacronym{cn}{CN}{Complex Network}
\newacronym{automl}{AutoML}{Automated Machine Learning}

\usetikzlibrary{decorations.pathmorphing}

\definecolor{cadmiumgreen}{rgb}{0.0, 0.42, 0.24}
\definecolor{cadmiumred}{rgb}{0.89, 0.0, 0.13}

\newcommand*\rfrac[2]{{}^{#1}\!/_{#2}}

\journal{Information Sciences}

\begin{document}

\begin{frontmatter}

\title{A meta-learning recommender system for hyperparameter tuning: predicting when tuning improves SVM classifiers}

\author[uspAddress,utfpr]{Rafael G. Mantovani\corref{mycorrespondingauthor}}
\ead{rafaelmantovani@utfpr.edu.br}
\cortext[mycorrespondingauthor]{Rafael Gomes Mantovani \\ Computer Engineering Department, Federal Technology University, Campus of Apucarana \\
R. Marc{\'i}lio Dias, 635 - Jardim Para{\'i}so, Apucarana - PR, Brazil, Postal Code 86812-460}

\author[unespAddress]{ Andr\'{e} L. D. Rossi}
\ead{alrossi@itapeva.unesp.br}

\author[uspAddress]{Edesio Alcoba\c{c}a}
\ead{edesio@usp.br}

\author[tueAddress]{\\Joaquin Vanschoren}
\ead{j.vanschoren@tue.nl}

\author[uspAddress]{Andr{\'e} C. P. L. F. {de Carvalho}}
\ead{andre@icmc.usp.br}

\address[uspAddress]{Institute of Mathematics and Computer Sciences, University of S\~ao Paulo, S\~ao Carlos - SP, Brazil}
\address[tueAddress]{ Eindhoven University of Technology, Eindhoven, Netherlands} 
\address[unespAddress]{Universidade Estadual Paulista, Campus de Itapeva, S\~{a}o Paulo, Brazil}

\address[utfpr]{Federal Technology University, Campus of Apucarana - PR, Brazil}
 

\begin{abstract}

For many machine learning algorithms, predictive performance is critically affected by the hyperparameter values used to train them. 
However, tuning these hyperparameters can come at a high computational cost, especially on larger datasets, while the tuned settings do not always significantly outperform the default values. 
This paper proposes a recommender system based on meta-learning to identify exactly when it is better to use default values and when to tune hyperparameters for each new dataset. 
Besides, an in-depth analysis is performed to understand what they take into account for their decisions, providing useful insights. An extensive analysis of different categories of meta-features, meta-learners, and setups across 156 datasets is performed. Results show that it is possible to accurately predict when tuning will significantly improve the performance of the induced models. The proposed system reduces the time spent on optimization processes, without reducing the predictive performance of the induced models (when compared with the ones obtained using tuned hyperparameters). We also explain the decision-making process of the meta-learners in terms of linear separability-based hypotheses. 
Although this analysis is focused on the tuning of Support Vector Machines, it can also be applied to other algorithms, as shown in experiments performed with decision trees.

\end{abstract}


\end{frontmatter}




\section{Introduction} 
\label{sec:intro}

Many \acrfull{ml} algorithms, among them \acrfullpl{svm}~\cite{vapnik:1995}, have been successfully used in a wide variety of problems. \acrshortpl{svm} are kernel-based algorithms that perform non-linear classification using a hyperspace transformation, i.e., they map data inputs into a high-dimensional feature space where the problem is possibly linearly separable. 
As most \acrshort{ml} algorithms, \acrshortpl{svm} are sensitive to their \acrfull{hp} values, which directly affect their predictive performance and depend on the data under analysis. The predictive performance of \acrshortpl{svm} is mostly affected by the values of four \acrshortpl{hp}: the kernel function ($k$), its width ($\gamma$) or polynomial degree ($d$), and the regularized constant ($C$).
Hence, finding suitable \acrshort{svm} \acrshortpl{hp} is a frequently studied problem~\cite{Horn:2016,Padierna:2017}. 
\acrshort{svm} \acrshort{hp} tuning is commonly modeled as a black-box optimization problem whose objective function is associated with the predictive performance of the SVM induced model.
Many optimization techniques have been proposed in the literature for this problem, varying from a simple \acrfull{gs} to the state of the art \acrfull{smbo} technique~\cite{Snoek:2012}. 
In \cite{Bergstra:2012}, Bergstra~\&~Bengio showed theoretically and empirically that \acrfull{rs} is a better alternative than \acrshort{gs} and is able to find good \acrshort{hp} settings when performing \acrshort{hp} tuning. Mantovani~et.~al.~\cite{Mantovani:2015a} also compared \acrshort{rs} with meta-heuristics to tune \acrshort{svm} \acrshortpl{hp}. A large amount of empirical experiments showed that \acrshort{rs} generates models with predictive performance as effective as those obtained by meta-heuristics.


However, regardless the optimization technique, hyperparameter tuning usually has a high computational cost, particularly for large datasets, with no guarantee that a model with high predictive performance will be obtained.
During the tuning, a large number of \acrshort{hp} settings usually need to be assessed before a ``good'' solution is found, requiring the induction of several models, multiplying the learning cost by the number of settings evaluated. Besides, several aspects, such as the complexity of a dataset, can influence the tuning cost.

When computational resources are limited, a commonly adopted alternative is to use the default \acrshort{hp} values suggested by \acrshort{ml} tools.
Previous works have pointed out that for some datasets, \acrshort{hp} tuning of \acrshortpl{svm} is not necessary~\cite{Ridd:2014}. Using default values largely reduce the overall computational cost, but, depending on the dataset, can result in models whose predictive performance is significantly worse than models produced by using \acrshort{hp} tuning.
The ideal situation would be to recommend the best alternative, default or tuned \acrshort{hp} values for each new dataset.


In this paper, we propose a recommender system to predict, when applying \acrshortpl{svm} to a new dataset, whether it is better to perform \acrshort{hp} tuning or it is sufficient to use default \acrshort{hp} values. This system, based on \acrfull{mtl}~\cite{Brazdil:2009}, is able to reduce the overall cost of tuning without significant loss in predictive performance. 
Another important novelty in this study is a descriptive analysis of how the recommendation occurs.
Although the recommender system is proposed for the \acrshort{hp} tuning of \acrshortpl{svm}, it can also be used for other \acrshort{ml} algorithms. To illustrate this aspect, we present an example where the recommender system is used for \acrshort{hp} tuning of a \acrfull{dt} induction algorithm.


The proposed recommender system can also be categorized as an \acrfull{automl} solution~\cite{Feurer:2015b}, since it aims to relieve the user from the repetitive and time-consuming tuning task, automating the process through \acrshort{mtl}. The \acrshort{automl} area is relatively new, and there still many questions to be addressed. This fact, and the emerging attention it has attracted from important research groups~\cite{Feurer:2015b,Kotthoff:2016} and large companies~\footnote{Google Cloud AutoML - \url{https://cloud.google.com/automl/}}~\footnote{Microsoft Custom Vision - \url{https://www.customvision.ai}}, highlights the importance of new studies in this area. 
An essential aspect for the success of \acrshort{automl} systems is to provide an automatic and robust tuning system, which also emphasizes the relevance of the problem investigated in this paper.


In summary, the main contributions of this study are:
\begin{itemize}

    \item the development of a modular and extensible \acrshort{mtl} framework to predict when default \acrshort{hp} values provide accurate models, saving computational time that would be wasted on optimization with no significant improvement;
    
    \item a comparison of the effectiveness of different sets of meta-features and preprocessing methods for meta-learning, not previously investigated;
    
    \item reproducibility of the experiments and analyses: all the code and experimental results are available to reproduce experiments, analyses and allow further investigations\footnote{The code is available in \texttt{Github} repositories, while experimental results are available on \acrshort{openml}~\cite{Vanschoren:2014} study pages. These links are provided in Table~\ref{tab:repo} at Subsection~\ref{subsec:repo}.}.
    
\end{itemize}


It is important to mention that we considered the proposed framework for predictive tasks, in particular, supervised classification tasks using \acrshortpl{svm}. However, the issues investigated in this paper can be easily extended to other tasks (such as regression) and other \acrshort{ml} algorithms~\footnote{A note on the generalization of the proposal is presented in Section~\ref{sec:generalization}}.


This paper is structured as follows: Section~\ref{sec:meta-learning} presents the basic \acrshort{mtl} concepts used in our approach. Section \ref{sec:related} defines the \acrshort{hp} tuning problem and presents a concise survey of prior work combining \acrshortpl{svm} with \acrshort{mtl} in some way. The complete experimental methodology covered to obtain the results is presented in Section \ref{sec:methods}. Results are discussed in Section~\ref{sec:results} while final considerations and conclusions are presented in Section~\ref{sec:conclusions}.


\section{Background on Meta-Learning} \label{sec:meta-learning}

Several \acrshort{ml} algorithms have been proposed for prediction tasks. 
However, since each algorithm has its inductive bias, some of them can be more appropriate for a particular data set. When applying a \acrshort{ml} algorithm to a dataset,
a higher predictive performance can be obtained if an algorithm whose bias is more adequate to the dataset is used. The recommendation of the most adequate \acrshort{ml} algorithm for a new dataset is investigated in an research area known as \acrfull{mtl}~\cite{Brazdil:2009}.

\acrshort{mtl} has been largely used for algorithm selection \cite{Ali:2006}, and for ranking \cite{Soares:2006} and prediction \cite{Reif:2014} of predictive performance of \acrshort{ml} algorithms. It investigates how to learn from previous \acrshort{ml} experiments.
According to Brazdil~et.~al.~\cite{Brazdil:2009}, meta-learning can be used to improve the learning mechanism itself after each training process.
In \acrshort{mtl}, the process of using a learning algorithm to induce a model for a data set is called base-learning. At the meta-level, likely useful information extracted from this process (meta-features) are used to induce a meta-model.
This meta-model can recommend the most promising learning algorithm, a set of the $N$ best learning algorithms or a ranking of learning algorithms according to their estimate predictive performance for a new dataset.
The knowledge extracted during this process is called meta-knowledge.  
The meta-features extracted from each dataset is a critical aspect. They must be sufficient to describe the main aspects necessary to distinguish the predictive performance obtained by different learning algorithms when applied to this dataset. As a result, it should allow the induction of a meta-model with good predictive performance. According to \cite{Vilalta:2004} three different sets of measures can be applied to extract meta-features:

\begin{itemize}

    \item [(i)] \textit{Simple}, \textit{Statistical} and \textit{Information-theoretic} meta-features~
    \cite{Brazdil:1994B}: these consist of simple measures about the input dataset, such as the number of attributes, examples and classes, skewness, kurtosis and entropy. They are the most explored subset of meta-features in literature~\cite{Feurer:2015b,Gomes:2012,Miranda:2012,Reif:2012,Reif:2014,Soares:2004};
    
    \item [(ii)] \textit{Model-based} meta-features~\cite{Bensusan:2000}: these are a set of properties of a model induced by a \acrshort{ml} algorithm for the dataset at the hand. For instance, if a decision tree induction algorithm is applied to the dataset, statistics about nodes, leaves and branches can be used to describe the dataset. They have also been used frequently in literature~\cite{Reif:2012,Reif:2014};
    
    \item [(iii)]\textit{Landmarking}~\cite{Pfahringer:2000}: the predictive performance obtained by models induced by simple learning algorithms, called landmarkers, are used to characterize a dataset. These measures were explored in studies such as~\cite{Feurer:2015b,Reif:2014}. 
    
\end{itemize}

Recently, new sets of measures have been proposed and explored in literature:
\begin{itemize}

    \item [(iv)] \textit{Data complexity}~\cite{Ho:2002}: this is a set of measures which analyze the complexity of a problem considering the overlap in the attributes values, the separability of the classes, and geometry/topological properties. They have been explored in~\cite{Garcia:2016}; and

    \item [(v)] \textit{Complex networks}~\cite{Morais:2013}: measures based on complex network properties are extracted from a network built with the data instances. These measures can only be extracted from numerical data. Thus, preprocessing procedures are required for their extraction. They were explored in~\cite{Garcia:2016}.

\end{itemize}



\section{Meta-learning for Hyperparameter tuning}
\label{sec:related}

As previously mentioned, there is a large number of studies investigating the use of  \acrshort{mtl} to automate one or more steps in the application of \acrshort{ml} algorithms for data analysis tasks. These studies can be roughly grouped into the following approaches, according to what \acrshort{mtl} does:

\begin{itemize}
\item it recommends \acrshort{hp} settings;
\item it predicts training runtime;
\item it recommends initial values for \acrshort{hp} optimization;
\item it estimates predictive performance for an \acrshort{hp} setting;
\item it predicts \acrshort{hp} tuning improvement/necessity.
\end{itemize}

\noindent Table~\ref{tab:svm_related} summarizes a comprehensive list of studies that either embedded or used \acrfull{mtl} to cope with the \acrshort{svm} \acrshort{hp} tuning problem. Next, these works are described in more detail. 


\subsection{Recommendation of HP settings}

The first approach considered \acrshort{hp} settings as independent algorithm configurations and predicted the best setting based on characteristics of the dataset under analysis. In this approach, the \acrshort{hp} settings are predicted without actually evaluating the model on the new dataset~\cite{Soares:2004}. In Soares~et.~al.~\cite{Soares:2004} and Soares~\&~Brazdil~\cite{Soares:2006}, the authors predicted the width ($\gamma$) of the \acrshort{svm} Gaussian kernel for regression problems. A finite set of $\gamma$ values was investigated for $42$ regression problems and the predictive performance was assessed using $10$-fold \acrshort{cv} and the \acrfull{nmse} evaluation measure. The recommendation of $\gamma$ values for new datasets used a \acrfull{knn} meta-learner.

Ali~\&~Smith-Miles~\cite{Ali:2006} presented a similar study but selected one among five different \acrshort{svm} kernel functions for $112$ classification datasets. They assessed model predictive performance for different \acrshort{hp} settings using $10$-\acrshort{cv} procedure and the simple \acrfull{acc} measure. Miranda~\&~Prud{\^e}ncio~\cite{Miranda:2013} proposed another \acrshort{mtl} approach, called \acrfull{at}~\cite{Leite:2012}, to select the \acrshortpl{hp} $\gamma$ and the soft margin ($C$). Experiments performed on $60$ classification datasets assessed the settings using a single $10$-\acrshort{cv} and the \acrshort{acc} measure.

Lorena~et.~al.~\cite{Lorena:2018} proposed a set of complexity meta-features for regression problems. One of the case studies evaluated was the \acrshort{svm} \acrshort{hp} tuning problem. The authors generated a finite grid of $\gamma$, C and $\epsilon$ (margin of tolerance for regression \acrshortpl{svm}) values, assessing them with a single 10-fold \acrshort{cv} and \acrshort{nmse} measure, considering $39$ regression problems. The recommendation of \acrshortpl{hp} for new unseen datasets was performed by a \acrshort{knn} distance-based meta-learner.


\subsection{Prediction of Training Runtime}

Other works investigated the use of \acrshort{mtl} to estimate the training time of classification algorithms when induced by different \acrshort{hp} settings. In Reif~et.~al.~\cite{Reif:2011}, the authors predicted the training time for several classifiers, including \acrshortpl{svm}. They defined a discrete grid of $\gamma \times C$ \acrshort{hp} settings, assessing these settings on $123$ classification datasets considering the \acrfull{pmcc} and the \acrfull{nae} performance measures. In Priya~et.~al~\cite{Priya:2012}, the authors conducted a similar study but used a \acrfull{ga} to optimize parameters and perform meta-feature selection of six meta-learners. Experiments were carried out over $78$ classification datasets assessing \acrshort{hp} settings using a $5$-fold \acrshort{cv} and the \acrfull{mad} evaluation measure.


\subsection{Recommendation of initial values for HP optimization}

\acrshort{mtl} has also been used to speed up the optimization of \acrshort{hp} values for classification algorithms~\cite{Feurer:2015b,Gomes:2012,Miranda:2012,Reif:2012}.
In Gomes~et.~al.~\cite{Gomes:2012} \acrshort{mtl} is used to recommend \acrshort{hp} settings 
as initial search values by the \acrfull{pso} and \acrfull{ts} optimization techniques. Experiments were conducted in $40$ regression datasets adjusting the $C$ and $\gamma$ \acrshortpl{hp} to reduce the \acrshort{nmse} value. A
\acrshort{knn} meta-learner was used to recommend the initial search values.

Reif~et.~al.~\cite{Reif:2012} and Miranda~et.~al.~\cite{Miranda:2012} investigated, respectively, the use of \acrfullpl{ga} and different versions of \acrshortpl{pso} for the same task. In Miranda~et.al.~\cite{Miranda:2014}, the authors used multi-objective optimization to optimize the \acrshortpl{hp} to increase predictive the performance and the number of support vectors. These studies used simple accuracy measure and $10$-fold \acrshort{cv} to optimize $\gamma \times C$ \acrshort{hp} values.

The same approach is explored in a tool to automate the use of \acrfull{ml} algorithms, the \texttt{Auto-skLearn}~\cite{Feurer:2015b}. In this tool, \acrshort{mtl} is used to recommend \acrshort{hp} settings for the initial population of the \acrshort{smbo} optimization technique. 
The authors explored all the available \acrshort{svm} \acrshortpl{hp} in $140$ \acrshort{openml} classification datasets. It is the first and perhaps the only work that uses nested-\acrshortpl{cv} to assess \acrshort{hp} settings. Each setting was assessed in terms of the simple \acrshort{acc} measure.


\subsection{Estimation of predictive performance for an HP setting}

A more recent approach uses \acrshort{mtl} to estimate \acrshort{ml} algorithms' performance considering their \acrshortpl{hp}. In Reif~et.~al.~\cite{Reif:2014}, the authors evaluated different \acrshort{ml} algorithms in $54$ datasets, including \acrshortpl{svm}, and used the performance predictions to develop a \acrshort{mtl} system for automatic algorithm selection.

Wistuba~et.~al.~\cite{Wistuba:2018} adapted the acquisition function of surrogate models by one optimized meta-model. They evaluated several \acrshort{svm} \acrshort{hp} configurations in a holdout fashion procedure over $105$ datasets and used the meta-knowledge to predict the performance of new \acrshort{hp} settings for new datasets. The authors also proposed a new \acrfull{taf} that extended the original proposal by predicting the predictive performance of \acrshort{hp} settings for surrogate models.

Eggensperger~et.~al.~\cite{Eggensperger:2018} proposed a benchmarking approach of ``surrogate scenarios'', which extracts meta-knowledge from \acrshort{hp} optimization and algorithm configuration problems, and approximates the performance surface by regression models. One of the $11$ meta-datasets explored in the experimental setup has a set of \acrshortpl{svm} \acrshort{hp} settings assessed for the MNIST dataset. These settings were obtained executing a simple \acrshort{rs} method and three optimizers: \acrfull{roar}~\cite{Hutter:2009}, \acrfull{irace}~\cite{Lopez:2016}, and \acrfull{pils}.


\begin{sidewaystable}
\scriptsize
\centering
\caption{Summary of related studies applying to \acrshort{mtl} to \acrshortpl{svm}. Fields without information in the related study are marked with a hyphen.} 
\label{tab:svm_related}
\setlength{\tabcolsep}{2pt}
\begin{tabularx}{\textwidth}{lclccccccclcc}

    \toprule

    \multirow{3}{*}{\textbf{ Reference }} & \multirow{3}{*}{\textbf{ Year }}  & \multicolumn{1}{c}{\multirow{3}{*}{\textbf{ Meta-learning }}} & \multicolumn{4}{c}{\textbf{ \acrshort{svm}s }} & \multicolumn{1}{c}{\textbf{ Tuning }} & 
    \multicolumn{1}{c}{\textbf{ Meta }} 
    & \multicolumn{1}{c}{\textbf{ Number of }} & \multicolumn{1}{c}{\textbf{ Datasets' }} & \multicolumn{1}{c}{\textbf{ Evaluation }} & \multicolumn{1}{c}{\textbf{ Evaluation }}\\
    
    & & & \multicolumn{4}{c}{\textbf{ Parameters }} & \multicolumn{1}{c}{\textbf{ Techniques }} & 
    \multicolumn{1}{c}{\textbf{ Learner }} 
    & \multicolumn{1}{c}{\textbf{ Datasets }} & \multicolumn{1}{c}{\textbf{ Source }}
    & \multicolumn{1}{c}{\textbf{ Procedure }} & \multicolumn{1}{c}{\textbf{ Measure }}\\
    
    & & & \textbf{ $k$ } & \textbf{ $\gamma$ } & \textbf{ $C$ } & \textbf{ $d$ } \\
    
    \midrule
    
    \rule{0pt}{2ex}
    Soares~et~al.~\cite{Soares:2004} & 2004 & Recommends \acrshort{hp} settings & & $\bullet$ & & & \acrshort{gs} & \acrshort{knn} & 42 & UCI, METAL & 10-CV & NMSE \\
    \rule{0pt}{2ex}
    
    Soares~\&~Brazdil~\cite{Soares:2006} & 2006 & Recommends \acrshort{hp} settings & & $\bullet$ & & & \acrshort{gs} & \acrshort{knn} & 42 & UCI, METAL & 10-CV & NMSE \\
    \rule{0pt}{2ex}
    
    Ali~\&~Smith-Miles~\cite{Ali:2006} & 2006 & Recommends \acrshort{hp} settings & $\bullet$ & & & & \acrshort{gs} & C5.0 & 112 &  UCI, KDC & 10-CV & Acc \\
    \rule{0pt}{2ex}
    
    Miranda~\& & \multirow{2}{*}{2013} & \multirow{2}{*}{Recommends \acrshort{hp} settings} & & \multirow{2}{*}{$\bullet$} & \multirow{2}{*}{$\bullet$} & & \multirow{2}{*}{\acrshort{gs}} & \multirow{2}{*}{\acrshort{at}\:} & \multirow{2}{*}{60} & \multirow{2}{*}{UCI} & \multirow{2}{*}{10-CV} & \multirow{2}{*}{Acc} \\
    \hspace{0.1cm} ~Prud{\^e}ncio~\cite{Miranda:2013} \\
    \rule{0pt}{2ex}
    
    Lorena~\&~et~al.~\cite{Lorena:2018} & 2018 & Recommends \acrshort{hp} settings & & $\bullet$ & $\bullet$ & & \acrshort{gs} & \acrshort{knn} & 39 & UCI & 10-CV & \acrshort{nmse} \\
    \rule{0pt}{3ex}


    \multirow{2}{*}{Reif~et~al.~\cite{Reif:2011}} & \multirow{2}{*}{2011} & \multirow{2}{*}{Predicts training runtime} & & \multirow{2}{*}{$\bullet$} & \multirow{2}{*}{$\bullet$} & & \multirow{2}{*}{\acrshort{gs}} & \multirow{2}{*}{\acrshort{svm}} & \multirow{2}{*}{123} & \multirow{2}{*}{UCI} & \multirow{2}{*}{-} & PMCC \\ 
    & & & & & & & & & & & & NAE\\
    \rule{0pt}{2ex}

    \multirow{4}{*}{Priya~et~al.~\cite{Priya:2012}} & \multirow{4}{*}{2012} & \multirow{4}{*}{Predicts training runtime} & \multirow{4}{*}{$\bullet$} & \multirow{4}{*}{$\bullet$} & \multirow{4}{*}{$\bullet$} & & \multirow{4}{*}{\acrshort{ga}} & J48,\:\acrshort{svm} & \multirow{4}{*}{78} &  \multirow{4}{*}{UCI} & \multirow{4}{*}{5-CV} & \multirow{4}{*}{MAD} \\
    & & & & & & & & Bagging \\
    & & & & & & & & \acrshort{knn},\:NB \\
    & & & & & & & & Jrip \\
    \rule{0pt}{3ex}
    

    \multirow{2}{*}{Gomes~et~al.~\cite{Gomes:2012}} & \multirow{2}{*}{2012} &  Recommends initial values & & \multirow{2}{*}{$\bullet$} & \multirow{2}{*}{$\bullet$} & & \acrshort{gs} & \multirow{2}{*}{\acrshort{knn}} & \multirow{2}{*}{40} & \multirow{2}{*}{WEKA} & \multirow{2}{*}{10-CV} & \multirow{2}{*}{NMSE} \\
    & & for \acrshort{hp} optimization & & & & & \acrshort{pso},\acrshort{ts} \\
    \rule{0pt}{2ex} 
    
    \multirow{2}{*}{Miranda~et~al.~\cite{Miranda:2012}} & \multirow{2}{*}{2012} &  Recommends initial values & & \multirow{2}{*}{$\bullet$} & \multirow{2}{*}{$\bullet$} & & \multirow{2}{*}{\acrshort{gs}, \acrshort{pso}} & \multirow{2}{*}{\acrshort{knn}} &\multirow{2}{*}{40} & \multirow{2}{*}{UCI, WEKA} & \multirow{2}{*}{10-CV} & \multirow{2}{*}{Acc} \\
    & & for \acrshort{hp} optimization \\
    \rule{0pt}{2ex} 
    
    \multirow{2}{*}{Reif~et~al.~\cite{Reif:2012}} & \multirow{2}{*}{2012} & Recommends initial values & & \multirow{2}{*}{$\bullet$} & \multirow{2}{*}{$\bullet$} & & \multirow{2}{*}{\acrshort{gs},\acrshort{ga}} & \multirow{2}{*}{\acrshort{knn}} & \multirow{2}{*}{102} & \multirow{2}{*}{UCI, Statlib} & \multirow{2}{*}{10-CV} & \multirow{2}{*}{Acc} \\
     & & for \acrshort{hp} optimization\\
    \rule{0pt}{2ex}
        
    \multirow{2}{*}{Miranda~et~al.~\cite{Miranda:2014}} & \multirow{2}{*}{2014} &   Recommends initial values & & \multirow{2}{*}{$\bullet$} & \multirow{2}{*}{$\bullet$} & & \multirow{2}{*}{\acrshort{gs},\acrshort{pso}} & \multirow{2}{*}{\acrshort{knn}} & \multirow{2}{*}{100} &  \multirow{2}{*}{UCI} & \multirow{2}{*}{10-CV} & \multirow{2}{*}{Acc} \\
    & & for \acrshort{hp} optimization \\
    \rule{0pt}{2ex} 
    
    \multirow{2}{*}{Auto-skLearn~\cite{Feurer:2015b}} & \multirow{2}{*}{2015} & Recommends initial values for & \multirow{2}{*}{$\bullet$} & \multirow{2}{*}{$\bullet$} & \multirow{2}{*}{$\bullet$} & \multirow{2}{*}{$\bullet$} & \multirow{2}{*}{\:\acrshort{gs}, \acrshort{smbo}} & \multirow{2}{*}{\acrshort{knn}} & \multirow{2}{*}{140} & \multirow{2}{*}{OpenML} & \multirow{2}{*}{Nested-CV} & \multirow{2}{*}{Acc} \\
    & & for \acrshort{hp} optimization \\
    \rule{0pt}{3ex}
    

    \multirow{2}{*}{Reif~et~al.~\cite{Reif:2014}} & \multirow{2}{*}{2014} & Estimates predictive performance & & \multirow{2}{*}{$\bullet$} & \multirow{2}{*}{$\bullet$} & & \multirow{2}{*}{\acrshort{gs}} & \multirow{2}{*}{\acrshort{svm}} & \multirow{2}{*}{54} & \multirow{2}{*}{UCI, Statlib} & \multirow{2}{*}{10-CV} & RMSE \\
    & & for a \acrshort{hp} setting & & & & & & & & & & PMCC \\
    \rule{0pt}{2ex}

    \multirow{2}{*}{Wistuba~et~al.~\cite{Wistuba:2018}} & \multirow{2}{*}{2018} & Estimates predictive performance & \multirow{2}{*}{$\bullet$} & \multirow{2}{*}{$\bullet$} & \multirow{2}{*}{$\bullet$} & \multirow{2}{*}{$\bullet$} & \multirow{2}{*}{\acrshort{gs}} & \multirow{2}{*}{\acrshort{gp}} & \multirow{2}{*}{109} & \multirow{2}{*}{UCI, WEKA} & \multirow{2}{*}{Holdout} & - \\
    & & for a \acrshort{hp} setting \\

    \rule{0pt}{2ex}
    
    \multirow{2}{*}{Eggensperger~et~al.~\cite{Eggensperger:2018}} & \multirow{2}{*}{2018} & Estimates predictive performance & \multirow{2}{*}{$\bullet$} & \multirow{2}{*}{$\bullet$} & \multirow{2}{*}{$\bullet$} & \multirow{2}{*}{$\bullet$} & \acrshort{roar}, \acrshort{irace} & \multirow{2}{*}{\acrshort{rf}} & \multirow{2}{*}{11} & \multirow{2}{*}{AClib} & \multirow{2}{*}{Holdout} & RMSE \\
    & & for a \acrshort{hp} setting & & & & & \acrshort{rs}, \acrshort{pils} & & & & & \acrshort{scrr} \\
    \rule{0pt}{2ex}
    
    Ridd~\& & \multirow{2}{*}{2014} & \multirow{2}{*}{Predicts \acrshort{hp} tuning improvement} & & \multirow{2}{*}{$\bullet$} & \multirow{2}{*}{$\bullet$} & & \multirow{2}{*}{\acrshort{pso}} & J48, RF & \multirow{2}{*}{326} & \multirow{2}{*}{UCI, WEKA} & \multirow{2}{*}{-} & \multirow{2}{*}{AUC} \\
    \hspace{0.1cm} Giraud-Carrier~\cite{Ridd:2014} & & & & & & & & \acrshort{svm} \\
    \rule{0pt}{2ex}
    
    \multirow{2}{*}{Mantovani~et~al.~\cite{Mantovani:2015b}} & \multirow{2}{*}{2015} & \multirow{2}{*}{Predicts \acrshort{hp} tuning necessity} & & \multirow{2}{*}{$\bullet$} & \multirow{2}{*}{$\bullet$} & & \acrshort{gs}, \acrshort{rs}, \acrshort{eda}
    &  J48, SVM, LR & \multirow{2}{*}{143}& \multirow{2}{*}{UCI} &  \multirow{2}{*}{Nested-CV}  & \multirow{2}{*}{BAC} \\
    & & & & & & & \acrshort{ga},\acrshort{pso} & NB, \acrshort{knn}, RF \\
    \rule{0pt}{3ex}

    Sanders~\& & \multirow{2}{*}{2017} & \multirow{2}{*}{Predicts \acrshort{hp} tuning improvement} & \multirow{2}{*}{$\bullet$} & \multirow{2}{*}{$\bullet$} & \multirow{2}{*}{$\bullet$} & \multirow{2}{*}{$\bullet$} & \multirow{2}{*}{\acrshort{ga}} & \multirow{2}{*}{\acrshort{mlp}} & \multirow{2}{*}{229} & \multirow{2}{*}{OpenML} & \multirow{2}{*}{10-CV} & \multirow{2}{*}{AUC} \\
    \hspace{0.1cm} Giraud-Carrier~\cite{Sanders:2017} & & & & & & & & \\

    \bottomrule
    \end{tabularx}
\end{sidewaystable}


\subsection{Prediction of HP tuning improvement/necessity}
\label{subsec:tuning_necessity}

Although the studies mentioned in this section are the most related to our current work regarding the proposed modeling, they have different goals. While Ridd~\&~Giraud-Carrier ~\cite{Ridd:2014} and Sanders~\&~Giraud-Carrier~\cite{Sanders:2017} are concerned with predicting tuning improvement, Mantovani~et~al.~\cite{Mantovani:2015b} and the present study aimed to predict when \acrshort{hp} tuning is necessary.

Ridd~\&~Giraud-Carrier~\cite{Ridd:2014} investigated a~\acrfull{cash} problem. They carried out experiments using \acrshort{pso} technique to search the hyperspace of this~\acrshort{cash} problem in $326$ binary classification datasets. Their \acrshort{mtl}-based method predicts whether \acrshort{hp} tuning would lead to a considerable increase in accuracy considering a pool of algorithms, including \acrshort{svm}.Even though this is one of the first studies in this direction, we could point out some drawbacks:
\begin{itemize}
    \item the proposed method does not identify which algorithm and correspondent \acrshort{hp} values the user should run to achieve an improved performance;
    \item there is no guarantee that training and testing data are not mutually exclusive;
    \item the rule to label the meta-examples  is  defined empirically, based on thresholds of the difference of the accuracy between default and tuned HP values;
    \item all the datasets are binary classification problems; and
    \item it is not possible to reproduce the experiments, specially base-level tuning since most of the details are not explained, and the code is not available.
\end{itemize}

Sanders~\&~Giraud-Carrier~\cite{Sanders:2017} used a \acrshort{ga} technique for \acrshort{hp} tuning of three different \acrshort{ml} algorithms, including~\acrshortpl{svm}. Their experimental results with $229$ ~\acrshort{openml} classification datasets showed that tuning almost always yielded significant improvements compared to default \acrshort{hp} values. 
Thus, they focused on the regression task of predicting how much improvement can be expected by tuning \acrshort{hp} compared to default values. They also addressed this task using \acrshort{mtl}. However, their study presents some limitations, such as:
\begin{itemize}
    \item the optimization process of \acrshort{svm} hyperparameters were computationally costly and did not finish for most of the datasets;
    \item the meta-learner was not able to predict hyperparameter tuning improvements for \acrshort{svm} in those datasets whose tuning process finished;
    \item there is no guarantee that the generated meta-examples are different from each other (intersection between training and test data), since OpenML stores different versions of the same dataset. This could lead to biased results; and
    \item experiments are not reproducible since most of the details are not explained, and the code is not available.
\end{itemize}

Mantovani~et~al.~\cite{Mantovani:2015b} proposed a~\acrshort{mtl} recommender system to predict when~\acrshort{svm}~\acrshort{hp} tuning is necessary, i.e., when tuning is likely to improve the generalization power of the models. The meta-dataset was created by extracting characteristics based on simple and data complexity measures from $143$ classification datasets. In the base-level, different meta-heuristics (PSO, GA and EDA) were used to tune the SVM \acrshortpl{hp} using a nested-\acrshort{cv} resampling strategy. An ensemble of meta-models achieved the best predictive performance assessed by the F-Score using simple meta-features. Besides these promising results, this study presents some shortcomings, such as:
\begin{itemize}
    \item the best predictive performance at the meta-level is moderately low; 
    \item when the method recommends tuning, the meta-heuristic which would lead to the best performance is not recommended;
    \item only two default \acrshort{hp} settings were investigated. In general, users try more than two settings before tuning;
    \item there is no evidence that this method and the results can be generalized  to other \acrshort{ml} algorithms.
\end{itemize}

The main differences between the proposed approach and the most related work are shown in Table \ref{tab:most_related_work}. It is important to note that although these are the most similar studies we have found in the literature, they addressed different problems. Furthermore, the meta-datasets generated by each study were also different, since they were generated using different datasets, target algorithms, meta-features, and labeling rules.
Because of these particularities, the straightforward comparison of these studies is unfeasible. The only free choice we could explore is the same meta-features adopted by them. In fact, Ridd~\&~Giraud-Carrier~\cite{Ridd:2014} and Sanders~\&~Giraud-Carrier~\cite{Sanders:2017} used a total of $68$ meta-features which are included in our experimental setup (Section~\ref{subsec:metafeatures}).

Based on the literature, we realized that there is room for improvement in terms of predicting \acrshort{hp} tuning necessity for \acrshort{ml} algorithms and to better understand this meta-learning process. Our present work attempts to fill this gap yielding meta-models with high predictive performance and reasons why their decisions were made. To do this, we have comprehensive and systematically evaluated different categories of meta-features and preprocessing tasks, such as meta-feature selection and data balancing, and different default \acrshort{hp} values.


\begin{table*}[htb]
    \centering
    \caption{The most related studies to our current approach. In the Goal prediction column, ``improv.'' means improvement prediction. In the Task column, ``class'' denotes  classification, while ``regr'' denotes regression.}
    \label{tab:most_related_work}
    
    \setlength{\tabcolsep}{3pt}
    \begin{tabularx}{\textwidth}{llclccl}
        \toprule
        
        \multicolumn{1}{c}{\multirow{2}{*}{\textbf{Study}}} & \multicolumn{2}{c}{\textbf{Goal}} & \textbf{ Tuning setup } & \textbf{ Labeling } & \textbf{ Target } & \textbf{Meta} \\ 
        & \textbf{prediction} & \textbf{Task} & \textbf{ (base-level) } & \textbf{ rule } & \textbf{ algorithm }  & \textbf{features} \\
        \midrule
        
        \rule{0pt}{1ex}
         & \multirow{4}{*}{~improv.} & \multirow{4}{*}{class} & \multirow{4}{*}{Not detailed} & & & Simple \\
        Ridd~\& & & & & Accuracy & CASH & Statistical \\
        \hspace{0.1cm} Giraud-Carrier~\cite{Ridd:2014} & & & & threshold & (20 algs.) & Landmarkers \\
        & & & & & & Model-based \\
        \midrule
        
        \rule{0pt}{3ex}
        & & \multirow{4}{*}{class~} & Nested-CVs & & \multirow{4}{*}{SVM} & \\
        Mantovani~& tuning & & holdout (inner) & Confidence & & Simple\\
        \hspace{0.1cm}~et~al.~\cite{Mantovani:2015b} & necessity & & 10-CV (outer) & interval & & Data complexity \\
        & & & BAC (fitness) \\
        \midrule

        \rule{0pt}{3ex}
         & \multirow{4}{*}{~improv.} & \multirow{4}{*}{regr~} & & & CART & Simple\\
        Sanders~\& & & & 10-CV (single) & Confidence & MLP & Statistical\\
        \hspace{0.1cm}Giraud-Carrier~\cite{Sanders:2017} & & & AUC (fitness) & interval & SVM & Landmarkers \\
        & & & & & & Model-based \\
        \midrule
       
        \rule{0pt}{3ex}
         \multirow{4}{*}{Current study} & & \multirow{4}{*}{class} & Nested-CVs & & & \\
        & tuning & & 3-fold (inner) & Wilcoxon & SVM & Many \\
        & necessity & & 10-fold (outer) & test & J48 & (see Section~\ref{subsec:metafeatures})\\ 
        & & & BAC (fitness) \\
        \bottomrule
    \end{tabularx}
\end{table*}


\subsection{Summary of Literature Overview}

The literature review carried out by the authors found a large increase in the use of \acrshort{mtl} for tasks related to \acrshort{svm} \acrshort{hp} tuning. 
The authors found $18$ related works, but only three of them investigated specifically when \acrshort{hp} tuning is necessary or its improvement~(see Section~\ref{subsec:tuning_necessity}). 
Overall, the following aspects were observed:

\begin{itemize}
    
    \item fourteen of the studies created the meta-knowledge using \acrshort{gs} to tune the $\gamma \times C$ \acrshort{hp};
    
    \item most of the studies also evaluated the resultant models with a single \acrshort{cv} procedure and the simple \acrshort{acc} evaluation measure;
    
    \item half of the studies used in most $100$ datasets. In~\cite{Ridd:2014}, the authors used more than $300$ datasets, but all of them for binary classification;
    
    \item all investigated a small number of categories to generate meta-features;

    \item nine of the studies used only \acrshort{knn} as meta-learner;

    \item three of the studies applied meta-feature selection techniques to the meta-features;
    
    \item two of the studies provided the complete resources necessary for the reproducibility of experiments;
    
    \item None of the studies combined all these six previous issues.
    
\end{itemize}

In order to provide new insights in the investigation of how the use of \acrshort{mtl} in the \acrshort{svm} \acrshort{hp} tuning process can affect its predictive performance, this paper extends previous works by exploring:

\begin{itemize}
\item Meta-features produced by measures from different categories; 
\item Use of different learning algorithms as meta-learners;
\item Adoption of a reproducible and rigorous experimental methodology at base and meta-learning levels; and 
\item Assessment of the use of meta-feature selection techniques to evaluate and select meta-features.
\end{itemize}

One of the main contributions of this paper is the analysis of the meta-model predictions to identify when it is better to use default or tuned \acrshort{hp} values for the \acrshortpl{svm}, and which meta-features have a major role in this identification.



\section{Experimental methodology} 
\label{sec:methods}

In this paper, experiments were carried out using \acrshort{mtl} ideas to predict whether hyperparameter tuning can significantly improve \acrshort{svm} induced models, when compared with performance provided by their default hyperparameter values\footnote{The \texttt{e1071} package was used to implement \acrshortpl{svm}. It is the LibSVM~\cite{Chang:2011} interface to the R environment. }. The framework treats the recommendation problem as a binary classification task and is formally defined as follows:


Let $\mathcal{D}$ be the dataset collection. Each dataset $d_{j} \in \mathcal{D}$ is described by a vector $mf(d_j) = (mf_{j,1},...,mf_{j,K})$ of $K$ meta-features, with~$mf_{j,k} \in \mathcal{M}$, the set of all known meta-features.
Additionally, let $\Omega$ be a statistical labeling rule based on
the prior evaluations from tuned and default hyperparameter settings ($\mathcal{P}$).
Given a significance level $\alpha$, $\Omega$ maps prior performances to a binary classification task: $\Omega: P \times \alpha \rightarrow C\:|\: C = \{tune, not\_tune\}$. Thus, we can train a \emph{meta-learner} $\mathcal{L}$ to predict whether optimization will lead to significant improvement on new datasets $d_{i} \notin \mathcal{D}$,~i.e.:
\begin{equation}
  \mathcal{L}: \mathcal{M} \times \Omega \rightarrow C
\end{equation}

\begin{figure}[h]
    \centering
    \includegraphics[width=\textwidth]{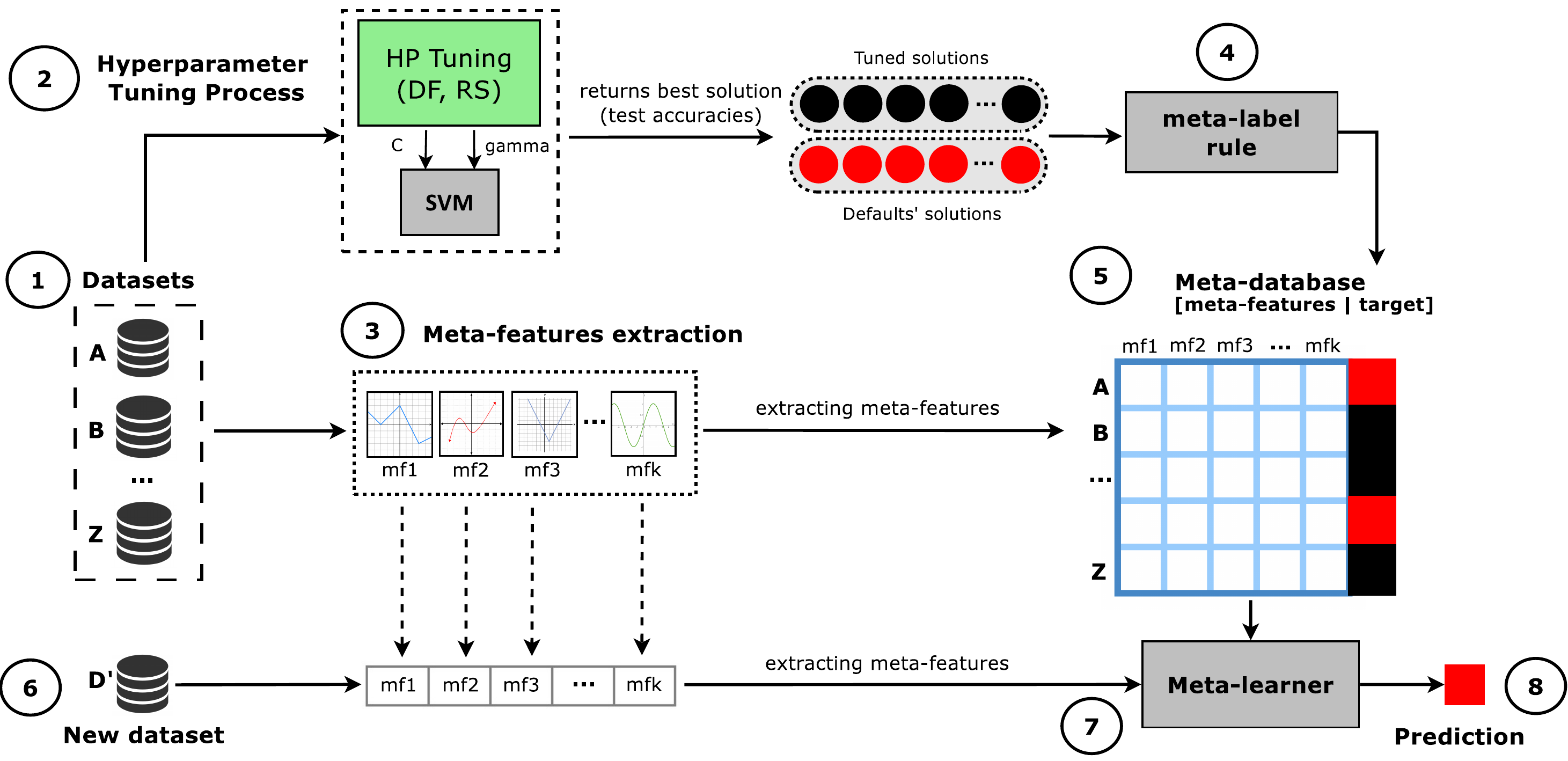}
    \caption{Meta-learning system to predict whether hyperparameter tuning is required (Adapted from~\cite{Mantovani:2015b}). At the figure, ''mf'' means meta-feature.}
    \label{fig:mtl-framework}
\end{figure}

Figure~\ref{fig:mtl-framework} shows graphically the general framework, linking two-level learning steps: the \textit{base level}, where the hyperparameter tuning process is performed for different datasets ($\mathcal{D}$); and the \textit{meta level}, where the meta-features ($\mathcal{M}$) from these datasets are extracted, the meta-examples are labeled according to tuning experiments ($\Omega$, $\mathcal{P}$) and the recommendation to a new unseen dataset occurs ($d_i \notin \mathcal{D}$). Further subsections will describe in detail each one of its components.


\subsection{OpenML classification datasets}
\label{subsec:datasets}


The experiments used datasets from \acrshort{openml}~\cite{Vanschoren:2014}, a free scientific platform for standardization of experiments and sharing empirical results. 
\acrshort{openml} supports reproducibility since any researcher can have access and use the same data for benchmark purposes.
A total of $156$ binary and multiclass classification datasets ($\mathcal{D}$) from different 
application domains were selected for the experiments (Item 01 in Figure~\ref{fig:mtl-framework}). From all the available and active datasets, those meeting the following criteria were selected:
\begin{itemize}
    \item [(a)] number of features does not exceed $1,500$;
    \item [(b)] number of instances between $100$ and $50,000$;
    \item [(c)] must not be a reduced, modified or binarized version of the original classification problem\footnote{More details about dataset versions can be found in the \acrshort{openml} paper~\cite{Vanschoren:2014} and documentation page: \url{https://docs.openml.org/\#data}.};
    \item [(d)] must not be an adaptation of a regression dataset;
    \item [(e)] all the classes must have at least $10$ examples, enabling the use of stratified $10$-fold \acrshort{cv} resampling.
\end{itemize}
\noindent

\noindent
These criteria are meant to ensure a proper evaluation (a-b), e.g. datasets should not be so small or so large that they cause memory problems; they should not be too similar (c-d) (to avoid data leakage in our evaluation); and allow the use of 10-fold \acrshort{cv} stratified resampling, given the high probability of dealing with imbalanced datasets (e). We also excluded datasets already used in our related work on defining optimized defaults, resulting in 156 datasets to be used in our meta-dataset. All datasets meeting these criteria and their main characteristics are presented on the study page at OpenML\footnote{\url{https://www.openml.org/s/52/data}}.

In order to be suitable for \acrshortpl{svm}, datasets were preprocessed: any constant or identifier attributes were removed; the logical attributes were converted into values $\in \{0,1\}$; missing values were imputed by the median for numerical attributes, and a new category for categorical ones; all categorical attributes were converted into the 1-N encoding; all attributes were normalized with $\mu = 0$ and $\sigma=1$. The \texttt{OpenML}~\cite{OpenML:2017}\footnote{\url{https://github.com/openml/openml-r}} package was used to obtain and select datasets from the \texttt{\acrshort{openml}} website, while functions from the \texttt{mlr}~\cite{mlr:2016}\footnote{\url{https://github.com/mlr-org/mlr}} package were used to preprocess them. 


\subsection{SVM hyperparameter space}
\label{subsec:hp_space}

The \acrshort{svm} hyperparameter space used in the experiments is presented in Table~\ref{tab:hp-space}.
For each hyperparameter, the table shows its symbol, name, type, range/options, scale transformation applied, default values provided by LibSVM~\cite{Chang:2011} and whether it was tuned. Here, only the \acrfull{rbf} kernel is considered since it achieves good performances in general, may handle nonlinear decision boundaries, and has less numerical difficulties than other kernel functions (e.g., the values of the polynomial kernel may be infinite)~\cite{Hsu:2007}. For $C$ and $\gamma$, the selected range covers the hyperspace investigated in~\cite{Ridd:2014}. 
LibSVM default values are $C = 1$, and $\gamma = 1/N$, where $N$ is the number of features of the dataset under analysis\footnote{LibSVM default values can be consulted at \url{https://www.csie.ntu.edu.tw/~cjlin/libsvm/}}.


\begin{table}[h]
\begin{minipage}{\textwidth}
    \centering
    \caption{SVM hyperparameter space used in experiments. The following was shown for each hyperparameter:} its symbol, name, type, range/options, scale transformation applied, default values and whether it was tuned.
 
    \label{tab:hp-space}
        \begin{tabular}{cllcccc}
           \toprule
            \textbf{Symbol} & \textbf{ Hyperparameter } & \textbf{ Type } & \textbf{ Range/Options } & \textbf{ Scale } & \textbf{Default} & \textbf{Tuned} \\
            \midrule
            \rule{0pt}{1ex} 
            k & kernel & categorical &\{RBF\} & - & RBF & \textbf{x} \\
            \rule{0pt}{3ex} 
            C & cost & real & $[2^{-15}, 2^{15}]$ & log & 1 & \checkmark\\
            \rule{0pt}{3ex} 
            $\gamma$ & width of the kernel & real & $[2^{-15}, 2^{15}]$ & log &  $\rfrac{1}{N}$ & \checkmark \\
            \bottomrule
        \end{tabular}
\end{minipage}
\end{table}


\subsection{Hyperparameter tuning process}
\label{subsec:tuning}

The hyperparameter tuning process is depicted in Figure~\ref{fig:mtl-framework} (Item 2). Based on the defined hyperspace, \acrshortpl{svm} hyperparameters were adjusted through a \acrfull{rs} technique for all datasets selected.  
The tuning process was carried out using nested \acrshort{cv} resamplings~\cite{Krstajic:2014}, an ``\textit{unbiased performance evaluation methodology}'' that correctly accounts for any overfitting that may occur in the model selection (considering the hyperparameter tuning).
In fact, most of the important/current state of the art studies, including the \texttt{Auto-WEKA}\footnote{\url{http://www.cs.ubc.ca/labs/beta/Projects/autoweka/}}~\cite{Kotthoff:2016,Thornton:2013} and \texttt{Auto-skLearn}\footnote{\url{https://github.com/automl/auto-sklearn}}~\cite{Feurer:2015b} tools, have been using the nested \acrshort{cv} methodology for hyperparameter selection and assessment.
Thus, nested-\acrshort{cv}s were also adopted in this current study. The number of outer folds was defined as $M=10$ such as in~\cite{Krstajic:2014}. Due to runtime constraints, the number of inner folds was set to $N=3$. 


A budget with a maximum of $300$ evaluations per (inner) fold was considered. A comparative experiment using different budget sizes for \acrshortpl{svm} was presented in~\cite{Mantovani:2015a}. Results suggested that only a few iterations are required to reach good solutions in the optima hyperspace region. Indeed, in most of the cases, tuning has reached good performance values after $250$-$300$ steps. Among techniques used by the authors, the \acrfull{rs} was able to find near-optimum hyperparameter settings like the most complex tuning techniques did. Overall, they did not show statistical differences regarding performance and presented a runtime lower than population-based techniques\footnote{These findings go towards what was previously described in ~\cite{Bergstra:2012}.}.

Hence, the tuning setup detailed in Table~\ref{tab:hp_setup} generates a total of $90,000 = 10$ (outer folds)~$\times\:3$~(inner folds)~$\times\:300$~(budget)~$\times\:10$ (seeds) \acrshort{hp} settings during the search process for a single dataset. Tuning jobs were parallelized in a cluster facility provided by our university\footnote{\url{http://www.cemeai.icmc.usp.br/Euler/index.html}} and took four months to be completed.


\begin{table*}[htpb]
\centering
\caption{Hyperparameter base level learning experimental setup.}
\begin{tabular}
    {@{\extracolsep{\fill}}lll}
\toprule
\textbf{Element} & \textbf{Method} & \textbf{R package} \\
\midrule
\rule{0pt}{1ex} 
Tuning techniques & Random Search & \texttt{mlr} \\
\rule{0pt}{3ex} 
Base Algorithm & Support Vector Machines & \texttt{e1071} \\
\rule{0pt}{3ex} 
Outer resampling & 10-fold cross-validation & \texttt{mlr} \\
\rule{0pt}{3ex} 
Inner resampling & 3-fold cross-validation & \texttt{mlr} \\
\rule{0pt}{3ex} 
Optimized measure & $\{$Balanced per class accuracy$\}$ & \texttt{mlr} \\
\rule{0pt}{3ex} 
\multirow{2}{*}{Evaluation measure} & $\{$Balanced per class accuracy, & \multirow{2}{*}{\texttt{mlr}} \\
& Optimization paths $\}$ & \\
\rule{0pt}{3ex} 
Budget  & 300 iterations \\
\rule{0pt}{3ex} 
\multirow{2}{*}{Repetitions} & $10$ times with different seeds & - \\
& seeds = $\{1, \ldots, 10\}$ & - \\
\rule{0pt}{3ex} 
\multirow{2}{*}{Baselines} & LibSVM defaults & \texttt{e1071} \\
& optimized defaults \\

\bottomrule
 
\end{tabular}
\label{tab:hp_setup}
\end{table*}


\subsection{Meta-features} 
\label{subsec:metafeatures}

The meta-datasets used in the experiments were generated out of `\textit{meta-features}' ($\mathcal{M}$) describing each dataset (Figure~\ref{fig:mtl-framework} - Item 3). These meta-features were extracted by applying a set of measures $mf_{i}$ to the original datasets which obtain likely relevant characteristics from these datasets. 
A tool was developed to extract the meta-features and can be found on \texttt{GitHub}\footnote{https://github.com/rgmantovani/MfeatExtractor}, as presented in Table~\ref{tab:repo}. We extracted a set of $80$ meta-features from different categories, as described in Section~\ref{sec:meta-learning}. The set includes all the meta-features explored by the studies described in Subsection~\ref{subsec:tuning_necessity}. The exact number of meta-features used from each category can be seen in Table~\ref{tab:mf-categories}. A complete description of them may be found in Tables~\ref{tab:mfeats_1} and \ref{tab:mfeats_2} (\ref{app:features}).


\begin{table}[h!]
\caption{Meta-feature category used in experiments.}
\label{tab:mf-categories}
\centering
\begin{tabular}{llrl}

\toprule
\textbf{Acronym} & \textbf{Category} & \textbf{\#N} & \textbf{Description}  \\

\noalign{\smallskip}\hline\noalign{\smallskip}
\rule{0pt}{1ex} 
SM & Simple & 17  & Simple measures \\
\rule{0pt}{3ex} 
ST & Statistical & 7  & Statistical measures \\
\rule{0pt}{3ex} 
IN & Information-theoretic & 8 & Information theory measures\\
\rule{0pt}{3ex} 
MB & Model-based (trees) & 17 & Features extracted from decision tree models\\
\rule{0pt}{3ex} 
LM & Landmarking & 8 & The performance of some ML algorithms \\
\rule{0pt}{3ex} 
DC & Data Complexity & 14 & Measures that analyze the complexity of a problem\\
\rule{0pt}{3ex} 
CN & Complex Networks & 9 & Measures based on complex networks\\
\midrule
\rule{0pt}{1ex} 
& \textbf{Total} & 80 \\

\bottomrule

\end{tabular}
\end{table}


\subsection{Meta-targets}
\label{subsec:metadatasets}

The last meta-feature is the meta-target, whose value indicates whether the \acrshort{hp} tuning significantly improved the predictive performance of the \acrshort{svm} model, compared with the use of default values. 
Since the \acrshort{hp} tuning experiments contain several and diverse datasets, many of them may be imbalanced. Hence, the~\acrfull{bac} measure~\cite{Brodersen:2010} was used as the fitness value during tuning, as well as for the final model assessment at the base-level learning\footnote{These performance values are assessed by \acrshort{bac} using a nested-\acrshort{cv} resampling method.}. 

The so-called ``\textit{meta-label rule}'' $\Omega$ (Item 4 - Figure~\ref{fig:mtl-framework}) applies the Wilcoxon paired-test to compare the solutions achieved by the \acrshort{rs} technique ($\mathcal{P}_{tun}$) and the default \acrshort{hp} settings ($\mathcal{P}_{def}$). 
Given a dataset $d_i \in \mathcal{D}$ and a significance level ($\alpha$), if the \acrshort{hp} tuned solutions were significantly better than those provided by defaults, its corresponding meta-example is labeled as `\texttt{Tuning}' ($C_{tun}$); otherwise, it receives the `\texttt{Default}' label ($C_{def}$). 


When performing the Wilcoxon test, three different values of $\alpha=\{0.10,\:0.05,\:0.01\}$ were considered, resulting in three meta-datasets with different class distributions (Item 5 - Figure~\ref{fig:mtl-framework}).
The different significance levels ($\alpha$) influence how strict the recommending system is when evaluating if tuning improved models' performance compared to the use of default \acrshort{hp} values. The smaller the significance stricter it is, i.e., there must be greater confidence than the tuned hyper-parameter values obtained by improving the performance of the induced models.
It may also imply in different labels for the same meta-example when evaluating different $\alpha$ values.
The initial experimental designs only compared LibSVM suggested default values with the \acrshort{hp} tuned solutions. The resulting meta-datasets presented a high imbalance rate, prevailing the ``\texttt{Tuning}'' class. It was difficult to induce a meta-model with high predictive performance using this highly imbalanced data.
An alternative to deal with this problem was to consider the optimized default \acrshort{hp} values proposed in~\cite{Mantovani:2015c}. The optimized default values were obtained optimizing a common set of \acrshort{hp} values, able to induce models with high predictive performance, for a group of datasets.  


\begin{figure}
    \centering
    \includegraphics[width=\textwidth]{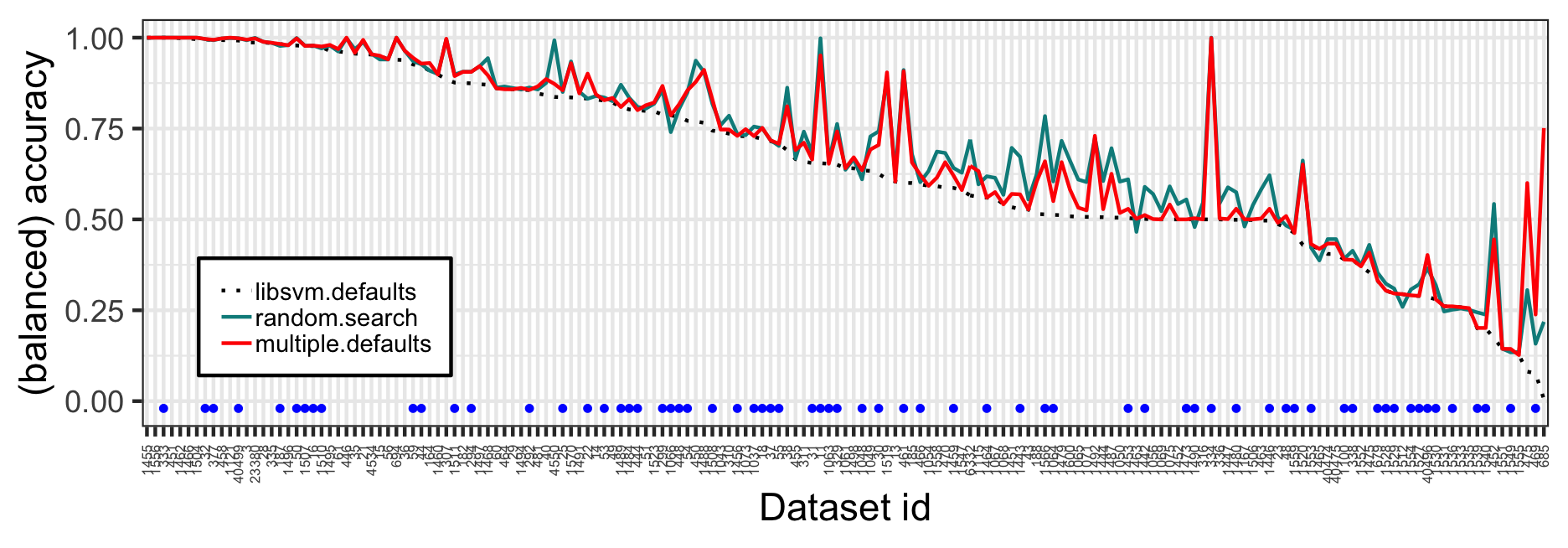}
    \caption{Average balanced per class accuracies comparing LibSVM default (libsvm.defaults), Multiple optimized default hyperparameter settings (multiple.defaults) and Random Search tuning technique (random.search) when defining the meta-target of each meta-example.}
    \label{fig:multiple_defaults}
\end{figure}


Figure~\ref{fig:multiple_defaults} illustrates the benefits of using multiple default settings: LibSVM and optimized default values. In this figure, the x-axis identifies datasets by their OpenML ids, listing them decreasingly by the balanced per class accuracy performances (y-axis) obtained using LibSVM defaults hyperparameter values. This figure shows three different curves:
\begin{itemize}
    \item \texttt{libsvm.defaults}: a black dotted line representing the averaged performance values obtained using LibSVM default hyperparameter values. It represents the choice of a user using LibSVM defaults;
    \item \texttt{random.search}: a green line representing the averaged performance values obtained using the \acrfull{rs} technique for tuning. It represents the choice of always tuning \acrshortpl{svm} hyperparameters; and
    \item \texttt{multiple.defaults}: a red line representing the best choice considering the LibSVM and optimized defaults hyperparameter values. It represents our approach, exploring multiple default values.
\end{itemize}

By looking at the difference between the black and green lines, it is possible to observe that tuned models using \acrshort{rs} outperformed models using default settings (provided by LibSVM) for around 2/3 of the datasets. However, when we consider multiple default settings (the best setting between LibSVM and optimized values), identified by the red curve, their performance values were close to the performance with tuned values. 
Thus, the meta-target labelling rule considered the difference between the predictive performances with tuned hyperparameters and the best predictive performances with multiple default \acrshort{hp} values.
A side effect of using multiple default \acrshort{hp} values is a more class-balanced meta-dataset, increasing the proportion of meta-examples labeled with ``default'' use.
As a result, the imbalance rate\footnote{imbalance rate = (majority class size / minority class size)} in the meta-datasets was reduced from $\approx 2.6$ to $1.7$.

Table~\ref{tab:meta_datasets} presents for each resultant meta-dataset: the $\alpha$ value used to generate the labels; the number of meta-examples, the number of meta-features and the class distribution. It is important to observe that none of these $156$ datasets were used in a related previous study that produced optimized default \acrshort{hp} setting~\cite{Mantovani:2015c}.

In our experimental setup, the null hypothesis of the statistical meta-label rule states that there is no significant difference between tuned and default \acrshort{svm} \acrshort{hp} settings. Since we are concerned about preventing tuning \acrshortpl{hp} when it is not necessary, a type I error is defined as labeling a meta-example as ``\texttt{Tuning}'' when its label is, in fact, ``\texttt{Defaults}'. Therefore, the lower the $\alpha$, the higher the probability that the improvement achieved by \textit{tuned} values is not due to chance. On the other hand, the higher the alpha, the lower the requirement that the performance gain by the tuning process is significant compared to default values.

Since we are controlling the error of labeling a meta-example as ``\texttt{Tuning}', smaller $\alpha$ values will lead to a greater number of ''\textit{default}” meta-examples. On the contrary, the greater the alpha value, the greater the number of meta-examples labeled as``\textit{tuning}". As can be seen in Table \ref{tab:meta_datasets}, a value of $\alpha=0.10$ implies more instances with the meta-target ``\texttt{Tuning}'' than when using $\alpha=0.01$.
In summary, if predictive performance is more critical, the user should set the significance level as high as possible (e.g., $\alpha=0.10$). On the other hand, if the user is concerned about computational cost, the significance level should be set to smaller values (e.g., $\alpha=0.01$). An example of this effect can be seen in Figure~\ref{fig:multiple_defaults}, where the blue dots represent all the datasets where defaults should be used, i.e., tuning is not statistically significant better (for $\alpha =0.05)$.


\begin{table}[h!]
    \caption{Meta-datasets generated from experiments with \acrshortpl{svm}.}
    \label{tab:meta_datasets}
    \centering
    \begin{tabular}{crcccc}
    
    \toprule
    \multirow{2}{*}{\textbf{Meta-dataset}} & \multirow{2}{*}{\textbf{$\alpha$}}
    & \textbf{Meta} & \textbf{Meta} & \multicolumn{2}{c}{\textbf{Class Distribution}} \\
    & & \textbf{examples} & \textbf{features} & \textit{Tuning} & \textit{Default} \\
    \midrule
    
    \rule{0pt}{1ex} 

    SVM\_90  & $0.10$ & 156 & 80 & 102 & 54 \\
    \rule{0pt}{3ex} 
    SVM\_95 & $0.05$ & 156 & 80 & 98 & 58 \\
    \rule{0pt}{3ex} 
    SVM\_99 & $0.01$ & 156 & 80 & 94 & 62 \\
    
    \bottomrule

    \end{tabular}
\end{table}


\subsection{Experimental Setup}
\label{subsec:setup}

Seven classification algorithms were used as meta-learners (Item 7 - Figure~\ref{fig:mtl-framework}): \acrfullpl{svm}, \acrfull{cart}, \acrfull{rf}, \acrfull{knn}, \acrfull{nb} \acrfull{lr} and \acrfullpl{gp}. These algorithms were chosen because they follow different learning paradigms with different learning biases. 
All seven algorithms were applied to the meta-datasets using a $10$-fold \acrshort{cv} resampling strategy and repeated $20$ times with different seeds (for reproducibility). 
All the meta-datasets presented in Table~\ref{tab:meta_datasets} are binary classification problems. Thus, meta-learners' predictions were assessed using the~\acrfull{auc} performance measure, a more robust metric than \acrshort{bac} for binary problems. Moreover, \acrshort{auc} also enable us to evaluate the influence of different threshold values on predictions. Three options were also investigated at the meta-level:
\begin{itemize}
    
    \item[(i)] \textit{Meta-feature Selection}: as each meta-example is described by many meta-features, it may be the case that just a small subset of them is necessary to induce meta-models with high accuracy. Thus, a \acrfull{sfs} feature selection option was added to the meta-learning experimental setup. The \acrshort{sfs} method starts from an empty set of meta-features, and in each step, the meta-feature increasing the performance measure the most is added to the model. It stops when a minimum required value of improvement ($alpha$=$0.01$) is not satisfied. Internally, it also performs a stratified $3$-fold \acrshort{cv} assessing the resultant models also according to the \acrshort{auc} measure;
    
    \item[(ii)] \textit{Tuning}: since the hyperparameter values of the meta-learners may also affect their performance, tuning of the meta-learners was also considered in the experimental setup. A simple \acrshort{rs} technique was performed with a budget of $300$ evaluations and resultant models assessed through an inner stratified $3$-fold \acrshort{cv} and \acrshort{auc} measure. Table~\ref{tab:app_space} (Appendix~\ref{app:mtl_space}) shows the hyperspace considered for tuning the meta-learners.
    
    \item[(iii)] \textit{Data balancing}: even using the optimized default \acrshort{hp} values, the classes in the meta-datasets were imbalanced.
    Thus, to reduce this imbalance, the \acrfull{smote}~\cite{Smote:2002} technique was used in the experiments.
    
\end{itemize}
\noindent Some of the algorithms' implementations selected as meta-learners use a data scaling process by default. This is the case of the \acrshort{svm}, \acrshort{knn} and \acrshort{gp} meta-learners. A preliminary experiment showed that removing this option decreases their predictive performance considerably, while it does not affect the other algorithms. When data scaling is considered for all algorithms, the performance values of \acrshort{rf}, \acrshort{cart}, \acrshort{nb} and \acrshort{lr} meta-learners were decreased. Thus, data scaling was not considered as an option, and the algorithms used their default procedures, with which they obtained their best performance values. 
Two baselines were also adopted for comparisons: a meta-dataset composed only by simple meta-features and another with data complexity ones.
Both categories of meta-features were investigated before by related studies listed in Section~\ref{subsec:tuning_necessity}.


\begin{table}[ht!]
\centering
\caption{Meta-learning experimental setup.}
\begin{tabular}
    {@{\extracolsep{\fill}}lll}

\toprule
\textbf{Element} & \textbf{Method} & \textbf{R package} \\
\midrule

  \rule{0pt}{1ex} 

\multirow{7}{*}{Meta-learner} & \acrfullpl{svm} & \texttt{e1071} \\
 & \acrfull{cart} & \texttt{rpart} \\
 & \acrfull{rf}   & \texttt{randomForest} \\
 & \acrfull{knn}  & \texttt{kknn} \\
 & \acrfull{nb}   & \texttt{e1071} \\ 
 & \acrfull{lr}   & \texttt{gbm} \\
 & \acrfullpl{gp} & \texttt{kernlab} \\
   \rule{0pt}{4ex} 
 
Resampling & $10$-fold \acrshort{cv} & \texttt{mlr} \\
  \rule{0pt}{4ex} 

\multirow{2}{*}{Meta-feature Selection} & Sequential Forward Search - $alpha=0.01$ & \multirow{2}{*}{\texttt{mlr}} \\
 & inner 3-CV - measure \acrshort{auc} & \\
 \rule{0pt}{4ex} 

\multirow{3}{*}{Tuning} & \acrfull{rs} & \multirow{3}{*}{\texttt{mlr}} \\
 & budget = 300 & \\
 & inner 3-CV - measure \acrshort{auc} & \\
 \rule{0pt}{4ex} 
 
\multirow{2}{*}{Data Balancing} & \acrshort{smote} & \multirow{2}{*}{\texttt{mlr}} \\
& oversampling rate = 2 & \\
\rule{0pt}{4ex} 
 
\multirow{2}{*}{Repetitions} & $30$ times with different seeds & - \\
& from the interval $\{1, \ldots, 30\}$ & - \\
\rule{0pt}{4ex} 

\multirow{2}{*}{Evaluation measures} & \acrshort{auc}& \multirow{2}{*}{\texttt{mlr}} \\
& predictions (prob) & \\
  \rule{0pt}{4ex} 

\multirow{2}{*}{Baselines} & Simple meta-features & - \\ 
& Data complexity meta-features & - \\

\bottomrule
 
\end{tabular}
\label{tab:mtl_setup}
\end{table}


\subsection{Repositories for the coding used in this study}
\label{subsec:repo}

Details of the base-level tuning and meta-learning experiments are publicly available in the \acrshort{openml} Studies (ids $52$ and $58$, respectively). In the corresponding pages, all datasets, classification tasks, algorithms/flows and results are listed and available for reproducibility. The code used for the \acrshort{hp} tuning process (\texttt{HpTuning}), extracting meta-features (\texttt{MfeatExtractor}), running meta-learning (\texttt{mtlSuite}), and performing the graphical analyses (\texttt{MtlAnalysis}) are hosted at \texttt{GitHub}. All of these repositories are also listed in Table~\ref{tab:repo}.

\begin{table}[ht!]
\centering
\caption{Repositories with tools developed by the authors and results generated by experiments.}
\begin{tabular}{ll}
        
    \toprule
    \textbf{Task/Experiment} & \textbf{Website/Repository} \\
    \midrule
    \rule{0pt}{3ex}
    Hyperparameter tuning code & \url{https://github.com/rgmantovani/HpTuning} \\
    \rule{0pt}{3ex}
    Hyperparameter tuning results & \url{https://www.openml.org/s/52} \\
    \rule{0pt}{3ex}
    Meta-feature extraction & \url{https://github.com/rgmantovani/MfeatExtractor} \\
    \rule{0pt}{3ex}
    Meta-learning code & \url{https://github.com/rgmantovani/mtlSuite} \\
    \rule{0pt}{3ex}
    Meta-learning results & \url{https://www.openml.org/s/58} \\
    \rule{0pt}{3ex}
    Graphical Analysis & \url{https://github.com/rgmantovani/MtlAnalysis} \\
    \bottomrule

\end{tabular}
\label{tab:repo}
\end{table}




\section{Results and Discussion} 
\label{sec:results}

The main experimental results are described in the next subsections. First, an overview of the predictive performance of the meta-models for the predicting task when it is worth performing \acrshort{svm} \acrshort{hp} tuning. Next, different experimental setups and preprocessing techniques, such as dimensionality reduction, are evaluated. Finally, the predictions and meta-knowledge produced by the meta-models are analyzed. 


\subsection{Average performance}
\label{sec:avg_perf}

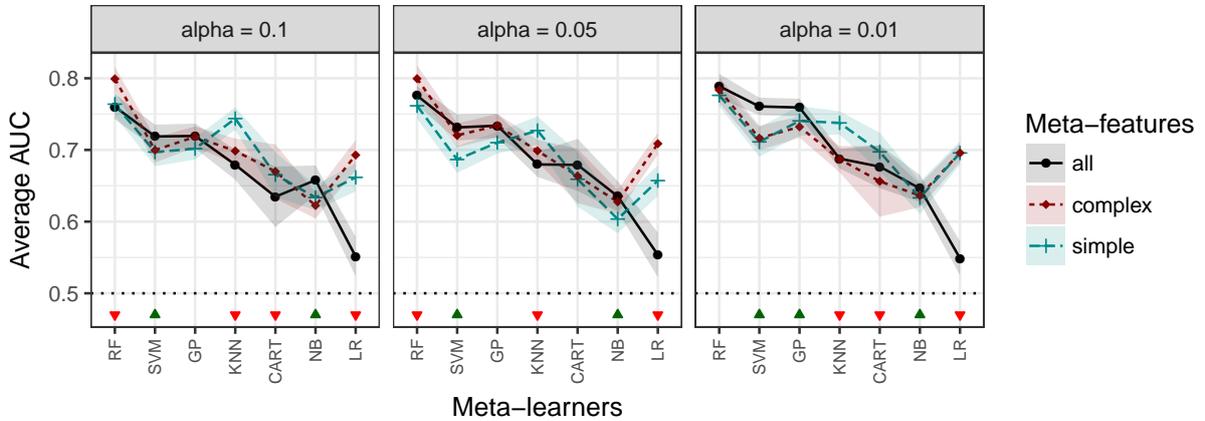
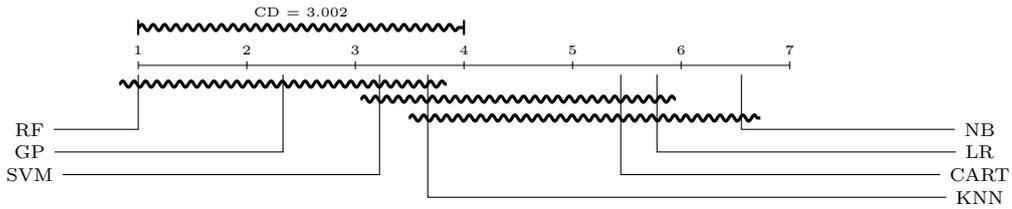
\begin{figure*}[htb!]
    \subfloat[Meta-learners average \acrshort{auc} performance on \acrshortpl{svm} meta-datasets. The black dotted line at $AUC=0.5$ represents the predictive performance of \texttt{ZeroR} and \texttt{Random} meta-models.]
    {
        \includegraphics
        [scale = 0.90] 
        {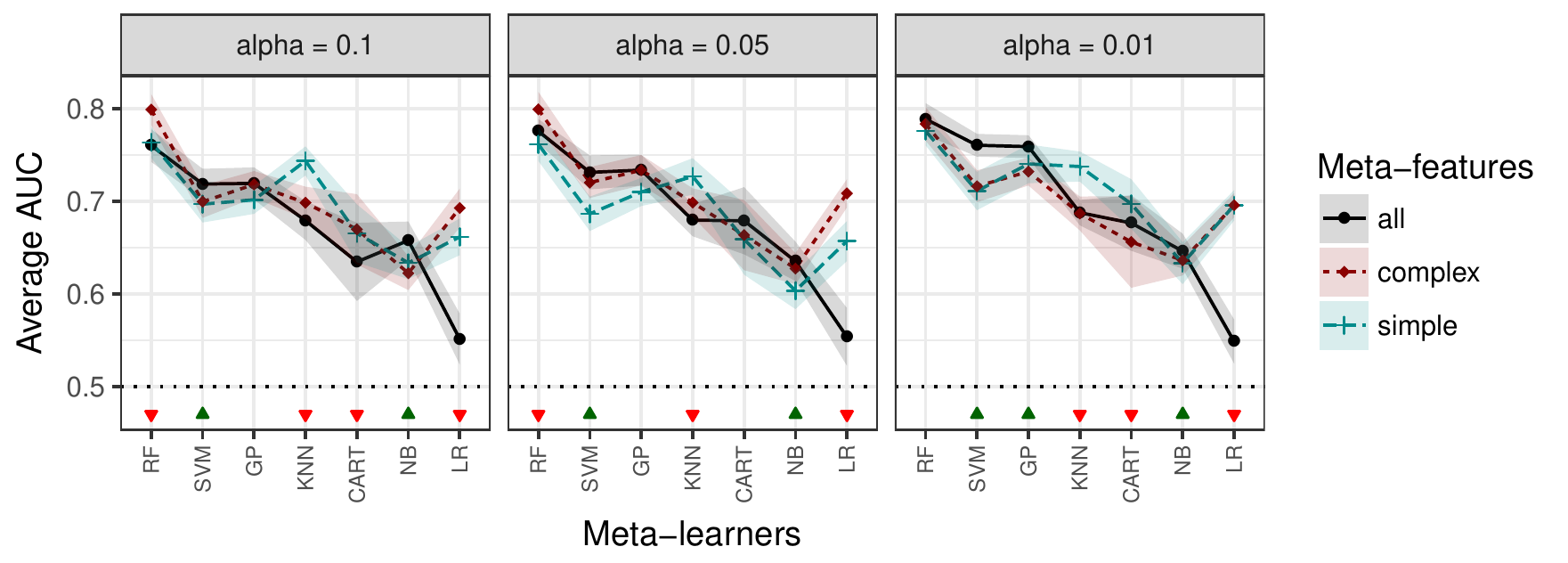}
        \label{fig:results}
    }
    \\
    \subfloat[
        Comparison of the \acrshort{auc} values of the induced meta-model according to the Friedman-Nemenyi test ($\alpha = 0.05)$. Groups of algorithms that are not significantly different are connected.
        ]
    {
        \centering 
        \begin{tikzpicture}[xscale=5]
        \node (Label) at  (00.7143,0.7) {\tiny{CD = 3.002}}; 
        \draw[decorate,decoration={snake,amplitude=.4mm,segment length=1.5mm,post length=0mm}, very thick, color = black](00.2857, 0.5) -- (01.1429, 0.5);
        \foreach \x in {00.2857,01.1429} \draw[thick,color = black] (\x, 0.4) -- (\x, 0.6);

        \draw[gray, thick](00.2857, 0) -- (02.0000, 0);
        \foreach \x in {00.2857,00.5714,00.8571,01.1429,01.4286,01.7143,02.0000}\draw (\x cm,1.5pt) -- (\x cm, -1.5pt);
        \node (Label) at (00.2857,0.2) {\tiny{1}};
        \node (Label) at (00.5714,0.2) {\tiny{2}};
        \node (Label) at (00.8571,0.2) {\tiny{3}};
        \node (Label) at (01.1429,0.2) {\tiny{4}};
        \node (Label) at (01.4286,0.2) {\tiny{5}};
        \node (Label) at (01.7143,0.2) {\tiny{6}};
        \node (Label) at (02.0000,0.2) {\tiny{7}};
        \draw[decorate,decoration={snake,amplitude=.4mm,segment length=1.5mm,post length=0mm}, very thick, color = black](00.2357,-00.2500) -- ( 01.0976,-00.2500);
        \draw[decorate,decoration={snake,amplitude=.4mm,segment length=1.5mm,post length=0mm}, very thick, color = black](00.8706,-00.4500) -- ( 01.7008,-00.4500);
        \draw[decorate,decoration={snake,amplitude=.4mm,segment length=1.5mm,post length=0mm}, very thick, color = black](00.9976,-00.7000) -- ( 01.9230,-00.7000);
        \node (Point) at (00.2857, 0){};  \node (Label) at (0.0,-00.8500){\scriptsize{RF}}; \draw (Point) |- (Label);
        \node (Point) at (00.6666, 0){};  \node (Label) at (0.0,-01.1500){\scriptsize{GP}}; \draw (Point) |- (Label);
        \node (Point) at (00.9206, 0){};  \node (Label) at (0.0,-01.4500){\scriptsize{SVM}}; \draw (Point) |- (Label);
        \node (Point) at (01.8730, 0){};  \node (Label) at (2.5,-00.8500){\scriptsize{NB}}; \draw (Point) |- (Label);
        \node (Point) at (01.6508, 0){};  \node (Label) at (2.5,-01.1500){\scriptsize{LR}}; \draw (Point) |- (Label);
        \node (Point) at (01.5555, 0){};  \node (Label) at (2.5,-01.4500){\scriptsize{CART}}; \draw (Point) |- (Label);
        \node (Point) at (01.0476, 0){};  \node (Label) at (2.5,-01.7500){\scriptsize{KNN}}; \draw (Point) |- (Label);
        \end{tikzpicture}
        \label{fig:results_friedman}
    }
    \centering 
    \caption{\acrshort{auc} performance values obtained by all meta-learners considering different meta-features' categories.~Results are averaged considering $30$ repetitions.}
    \label{fig:perf}
    
\end{figure*}

Figure~\ref{fig:perf} summarizes the predictive performance of different meta-learners for three different sets of meta-features, namely: \textit{all}, \textit{complex} and \textit{simple}. The former has all $80$ available meta-features, the \textit{complex} set contains only $14$ data complexity measures as meta-features and the latter consists of $17$ simple and general meta-features.

In Figure~\ref{fig:results}, the x-axis shows the meta-learners while the y-axis shows their predictive performance assessed by the \acrshort{auc} averaged over $30$ repetitions. In addition, it shows the impact of different \textit{alpha} ($\alpha$) levels for the Wilcoxon test for the definition of the meta-target labels.
The Wilcoxon paired-test with $\alpha=0.05$ was applied to assess the statistical significance of the predictive performance differences obtained by the meta-models with \textit{all} meta-features, when compared to the second best approach.

An upward green triangle (\textcolor{cadmiumgreen}{$\blacktriangle$}) at the x-axis identifies situations where using all the meta-features were statistically better. On the other hand, red downward triangles~(\textcolor{red}{$\blacktriangledown$}) show results where one of the alternative approaches was significantly better. In the remaining cases, the predictive performance of the meta-models were equivalents.


The best results were obtained by the \acrshort{rf} meta-learner using data complexity (\textit{complex}) meta-features, achieving \acrshort{auc} values nearly $0.80$ for all $\alpha$ levels. 
These meta-models were also statistically better than those obtained by other approaches at $\alpha=\{0.90,0.95\}$.
When $\alpha=0.99$, the \acrshort{rf} meta-learner using all the meta-features also generated a model with \acrshort{auc}~$\approx0.8$.

When the value of $\alpha$ in the meta-label rule is reduced, predictive performances using data complexity and all the available meta-features tend to show similar distributions. 
The meta-learners obtained their best \acrshort{auc} values with the highest assumption ($\alpha=0.99$). 
Overall, varying the $\alpha$ value did not substantially change  the predictive performance of the evaluated algorithms. 
In fact, few meta-examples had their meta-targets modified by the meta-rule with different values of $\alpha$. 
Thus, the predictions in the different scenarios are mostly the same and the performances remained similar.

Regarding predictive performance, \acrshort{rf}, \acrshort{svm}, \acrshort{gp} and \acrshort{knn} induced accurate meta-models for the three meta-dataset variations.
The \acrshort{auc} value varied in the interval $\{0.70, 0.80\}$. 
Even the \acrshort{lr}, depending on the meta-features used to represent the recommendation problem,  achieved reasonable \acrshort{auc} values. For comparison purposes, it is important to mention that both \texttt{Random} and \texttt{ZeroR}\footnote{This classifier simply predicts the majority class.} baselines obtained \acrshort{auc} of $0.5$ in all these meta-datasets\footnote{The \acrshort{auc} performance values were assessed using the implementations provided by the \texttt{mlr} R package.}.


The Friedman test~\cite{Demvsar:2006}, with a significance level of $\alpha = 0.05$, was used to assess the statistical significance of the meta-learners. 
In the comparisons, we considered the algorithms' performance across the combination of the meta-datasets and the categories of meta-features.
The null hypothesis states that all the meta-learners are equivalent regarding the \acrfull{auc} performance. When the null hypothesis is rejected, the Nemenyi post-hoc test is also applied to indicate when two different techniques are significantly different. 

Figure~\ref{fig:results_friedman} presents the resultant \acrfull{cd} diagram. Algorithms are connected when there are \textit{no} significant differences between them.
The top-ranked meta-learner was the \acrshort{rf} with an average rank of $1.0$, followed by \acrshort{gp} ($2.3$), \acrshort{svm} ($3.2$) and \acrshort{knn} ($3.6$). 
They did not present statistically differences among them, but mostly did when compared with simpler algorithms: \acrshort{cart} ($5.4$), \acrshort{lr} ($5.7$) and \acrshort{nb}($6.5$). Even not being statistically better than all the other choices, the \acrshort{rf} was always ranked at the top regardless of the meta-dataset and meta-features. 

Although the best result was obtained using \acrfull{dc} meta-features (``\texttt{complex}''), most of the meta-learners achieved their highest \acrshort{auc} performance values exploring all the available meta-features.
Thus, since we want to analyze the influence of different categories of meta-features when inducing meta-models, and given the possibility of selecting different subsets from all the categories, we decided to explore all of them in the next analysis.


\subsection{Evaluating different setups}

Due to the large difference among meta-learners results, three different setups were also evaluated to improve their predictive performances and enable a comprehensive analysis of the investigated alternatives:
\begin{itemize}
    \item [(i)] \texttt{featsel} - meta-feature selection via \acrfull{sfs}~\cite{mlr:2016};
    \item [(ii)] \texttt{tuned} - \acrshort{hp} tuning of the meta-learners using a simple \acrfull{rs} technique; 
    \item [(iii)] \texttt{smote}: dataset balancing with \acrshort{smote}~\cite{Smote:2002}.
\end{itemize}
\noindent They were compared with the original meta-data with no additional process (\texttt{none}), which is the baseline for these analyses.  
These setups were not performed at the same time to avoid overfitting, since the meta-datasets have $156$ meta-examples and, depending on their combinations, three levels of \acrshortpl{cv} would be used to assess models. 
For example: if feature meta-selection and \acrshort{hp} tuning were enabled at the meta-level, one \acrshort{cv} would be used for meta-feature selection, one for tuning and another to assess the resulting models.


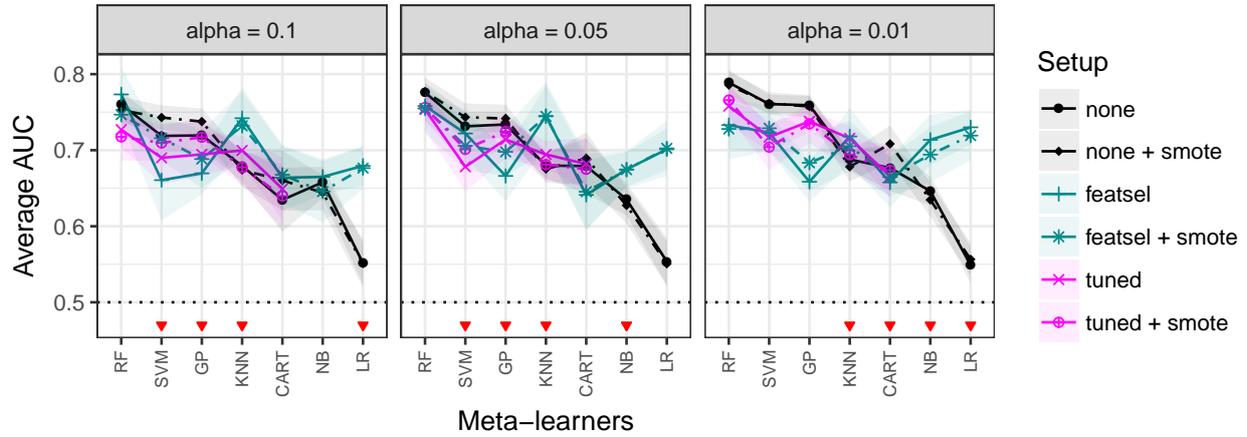
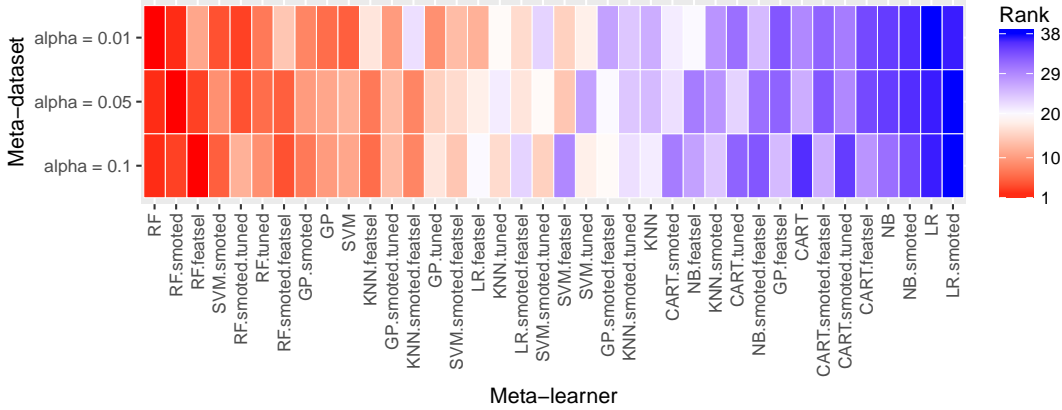
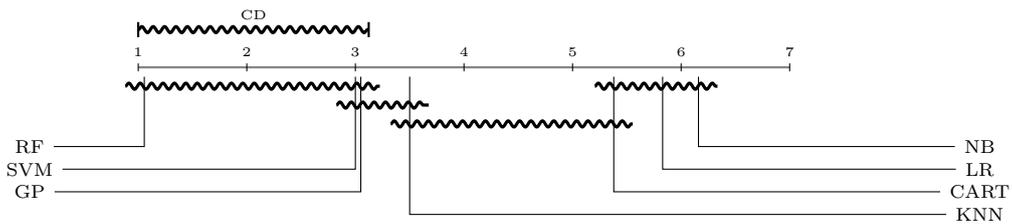
\begin{figure*}[ht!]

    \subfloat[Average \acrshort{auc} performance values.~The black dotted line at $AUC=0.5$ represents the predictive performance of \texttt{ZeroR} and \texttt{Random} meta-models.]
    {\includegraphics
    [scale = 0.93]
    {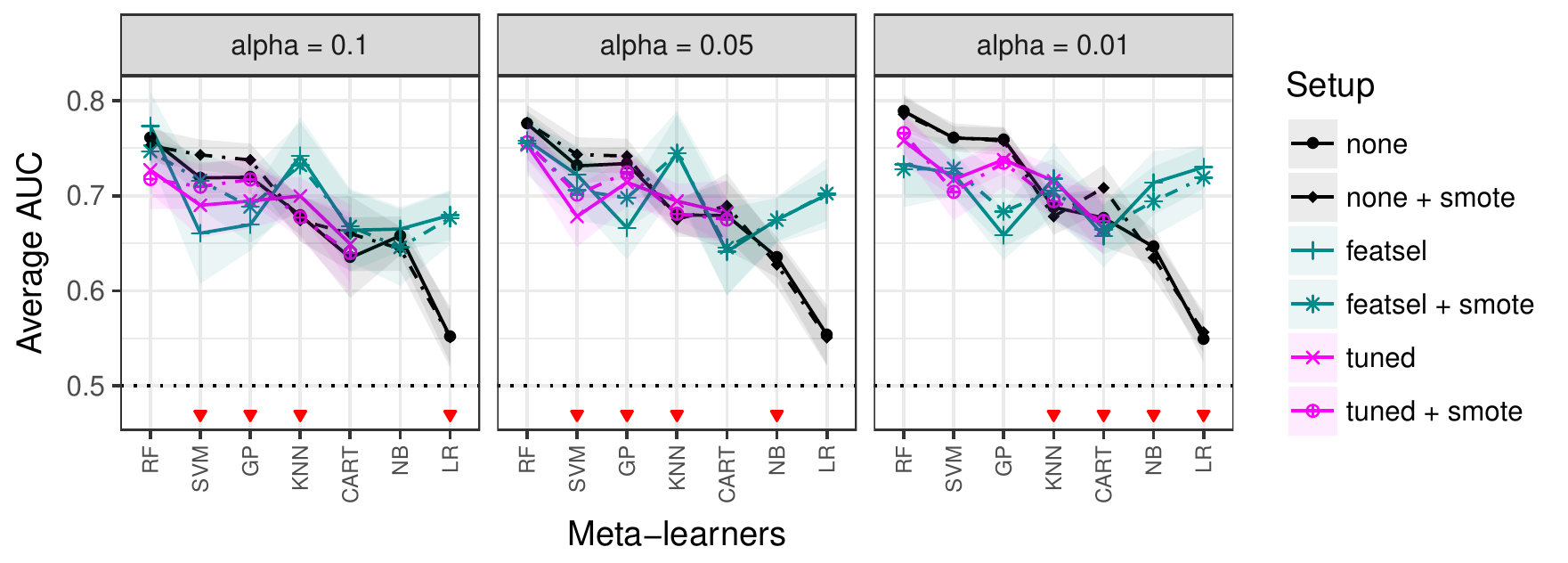}%
        \label{fig:auc_perf}
    }
    \\
    \subfloat[Average \acrshort{auc} ranking values.]
    {\includegraphics
    [scale = 0.73]
    {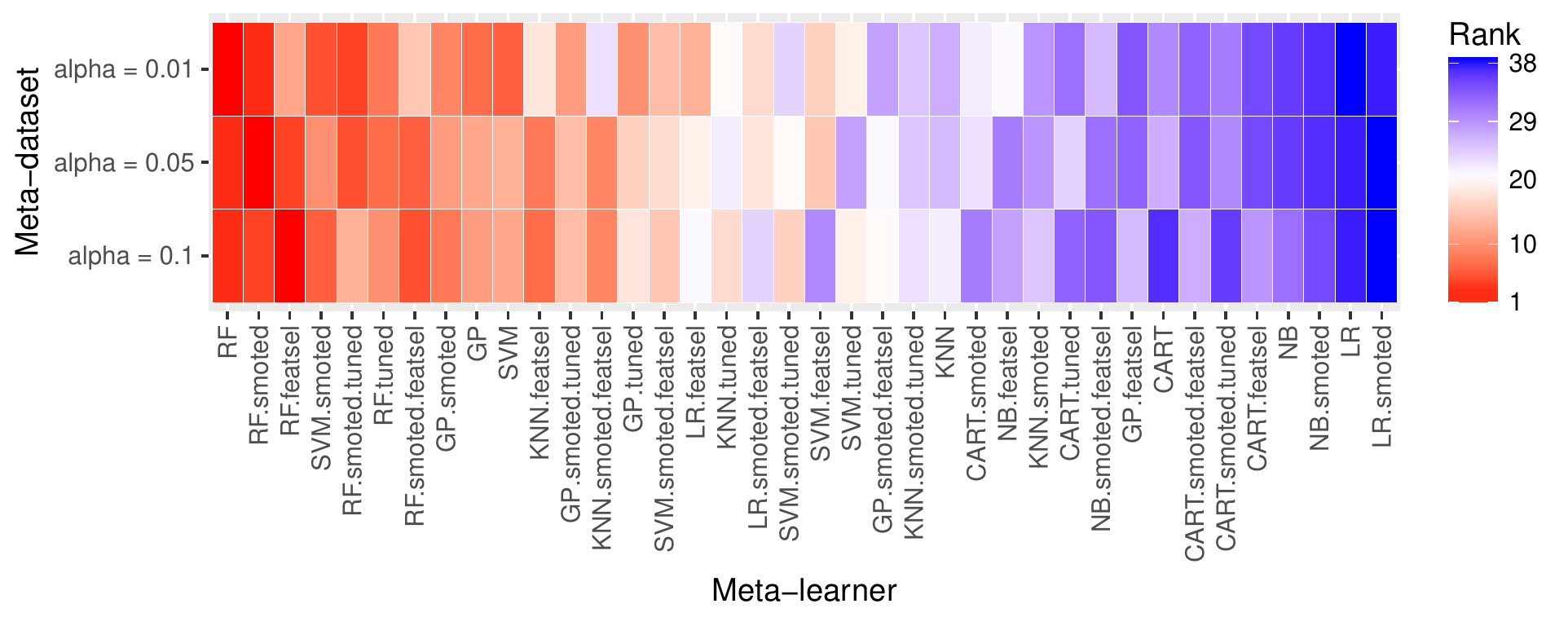}%
        \label{fig:auc_rk}
    }
    \\
    \subfloat[Comparison of the \acrshort{auc} values of the induced meta-model according to the Friedman-Nemenyi test ($\alpha = 0.05)$. Groups of algorithms that are not significantly different are connected.]
    {
        \centering 
        \begin{tikzpicture}[xscale=5]
            \node (Label) at  (00.5890,0.7) {\tiny{CD}};             \draw[decorate,decoration={snake,amplitude=.4mm,segment length=1.5mm,post length=0mm}, very thick, color = black](00.2857, 0.5) -- (00.8922, 0.5);
            \foreach \x in {00.2857,00.8922} \draw[thick,color = black] (\x, 0.4) -- (\x, 0.6);
                
            \draw[gray, thick](00.2857, 0) -- (02.0000, 0);
            \foreach \x in {00.2857,00.5714,00.8571,01.1429,01.4286,01.7143,02.0000}\draw (\x cm,1.5pt) -- (\x cm, -1.5pt);
            \node (Label) at (00.2857,0.2) {\tiny{1}};
            \node (Label) at (00.5714,0.2) {\tiny{2}};
            \node (Label) at (00.8571,0.2) {\tiny{3}};
            \node (Label) at (01.1429,0.2) {\tiny{4}};
            \node (Label) at (01.4286,0.2) {\tiny{5}};
            \node (Label) at (01.7143,0.2) {\tiny{6}};
            \node (Label) at (02.0000,0.2) {\tiny{7}};
            \draw[decorate,decoration={snake,amplitude=.4mm,segment length=1.5mm,post length=0mm}, very thick, color = black](00.2516,-00.2500) -- ( 00.9214,-00.2500);
            \draw[decorate,decoration={snake,amplitude=.4mm,segment length=1.5mm,post length=0mm}, very thick, color = black](00.8071,-00.5000) -- ( 01.0500,-00.5000);
            \draw[decorate,decoration={snake,amplitude=.4mm,segment length=1.5mm,post length=0mm}, very thick, color = black](00.9500,-00.7500) -- ( 01.5871,-00.7500);
            \draw[decorate,decoration={snake,amplitude=.4mm,segment length=1.5mm,post length=0mm}, very thick, color = black](01.4871,-00.2500) -- ( 01.8100,-00.2500);
            \node (Point) at (00.3016, 0){};  \node (Label) at (0.0,-01.0500){\scriptsize{RF}}; \draw (Point) |- (Label);
            \node (Point) at (00.8571, 0){};  \node (Label) at (0.0,-01.3500){\scriptsize{SVM}}; \draw (Point) |- (Label);
            \node (Point) at (00.8714, 0){};  \node (Label) at (0.0,-01.6500){\scriptsize{GP}}; \draw (Point) |- (Label);
            \node (Point) at (01.7600, 0){};  \node (Label) at (2.5,-01.0500){\scriptsize{NB}}; \draw (Point) |- (Label);
            \node (Point) at (01.6657, 0){};  \node (Label) at (2.5,-01.3500){\scriptsize{LR}}; \draw (Point) |- (Label);
            \node (Point) at (01.5371, 0){};  \node (Label) at (2.5,-01.6500){\scriptsize{CART}}; \draw (Point) |- (Label);
            \node (Point) at (01.0000, 0){};  \node (Label) at (2.5,-01.9500){\scriptsize{KNN}}; \draw (Point) |- (Label);
        \end{tikzpicture}
        \label{fig:cd_rks}
    }
    \centering 
    \caption{\acrshort{auc} performance values obtained by all meta-learners considering different experimental setups. Results are averaged over $30$ repetitions.}
    \label{fig:perf_setups}

\end{figure*}


Figure~\ref{fig:perf} summarizes the main aspects of these experimental results. 
The top figure shows the average \acrshort{auc} values for each experimental setup considering all the meta-learners and the $\alpha$ levels. 
The \acrshort{nb} and \acrshort{lr} meta-learners do not have any tunable \acrshort{hp}. 
Thus, their results in this figure are missing for the \texttt{tuned} setups (with and without \acrshort{smote}). 
Similarly to Figure~\ref{fig:results}, the statistical analysis is also presented. Every time an upward green triangle is placed at the x-axis, the raw meta-data (\texttt{none}) generated results statistically better than using the best of the experimental setups evaluated. On the other hand, red triangles indicate when tuning, meta-feature selection or \acrshort{smote} could statistically improve the predictive performance of the meta-models. In the remaining cases, the meta-models were equivalent.

Despite the different setups evaluated, \acrshort{rf} is still the best meta-learner for all $\alpha$ scenarios. 
It is followed by \acrshort{svm} and \acrshort{gp} versions using \acrshort{smote}. 
Depending on the experimental setup, \acrshort{knn} and \acrshort{lr} also presented good predictive performances. 
Regarding the \acrshort{hp} tuning (\texttt{tuned}) of the meta-learners, only for \acrshort{knn} the performances slightly improved for all the alpha values. Using just \acrshort{smote} resulted in improved results for \acrshort{svm}, \acrshort{gp} and \acrshort{cart} meta-learners. 
In general, it produced small improvements, but most of them were statistically significant. 
When used with tuning or meta-feature selection, it affected the algorithms in different ways: for \acrshort{svm} and \acrshort{gp}, the performance improved; for \acrshort{lr}, \acrshort{nb} and \acrshort{knn} there was no benefit; the other algorithms were not affected by its use.
The low gain obtained using \acrshort{smote} may be due to the fact that data imbalance was already reduced
using the optimized defaults when defining meta-targets.

Using meta-feature selection (\texttt{featsel}) deteriorated the performance of the \acrshort{svm}, \acrshort{rf}, \acrshortpl{gp} and \acrshort{cart} meta-learners. 
On the other hand, it clearly improved the \acrshort{knn}, \acrshort{lr} and \acrshort{nb} performances for most cases. 
\acrshort{knn} benefited from using a subset of meta-features to maximize the importance of more relevant meta-features.
For \acrshort{nb} and \acrshort{lr}, selecting a subset of the attributes reduced the presence of noise and irrelevant attributes.
Furthermore, it is important to observe that the meta-models induced with the selected features presented the highest standard deviation between the setups (light area along the curve). 
A possible reason is the different subsets selected every time meta-feature selection is performed for the $30$ repetitions. 


Additionally, Figure~\ref{fig:auc_rk}
presents a ranking with all the combinations of meta-learners and experimental setups.
At the x-axis, they are presented in ascending order according to their average ranking for the three scenarios ($\alpha$ values), shown at the y-axis. The more red the squares, the lower the ranking, i.e., the better the results.

As previously reported, \acrshort{rf} with no additional option was the best-ranked method, followed by its smoted versions. The \acrshort{svm}, \acrshort{gp} and \acrshort{knn} versions are in the next positions.
The Friedman test with a significance level of $\alpha=0.05$ was also used to assess the statistical significance of the meta-learners when using different experimental setups in different meta-datasets. Figure~\ref{fig:cd_rks} shows the resultant \acrshort{cd} diagram. Results are quite similar to those reported in Figure~\ref{fig:results_friedman}: the \acrshort{rf} algorithm was the best algorithm with an average ranking of $1.05$, and is statistically better than most of the meta-learners, except for \acrshort{svm} and \acrshort{gp}.


Since there were no improvements considering the maximum \acrshort{auc} values achieved so far,
and the results from the \acrshort{rf} meta-learners were still the best-ranked, 
the next subsections will analyze the relative importance of the meta-features according to the final \acrshort{rf} induced models.


\subsection{Importance of meta-features}
\label{subsec:rf}

\begin{figure}[h!]
    {
        \includegraphics 
        [width=\textwidth]
        {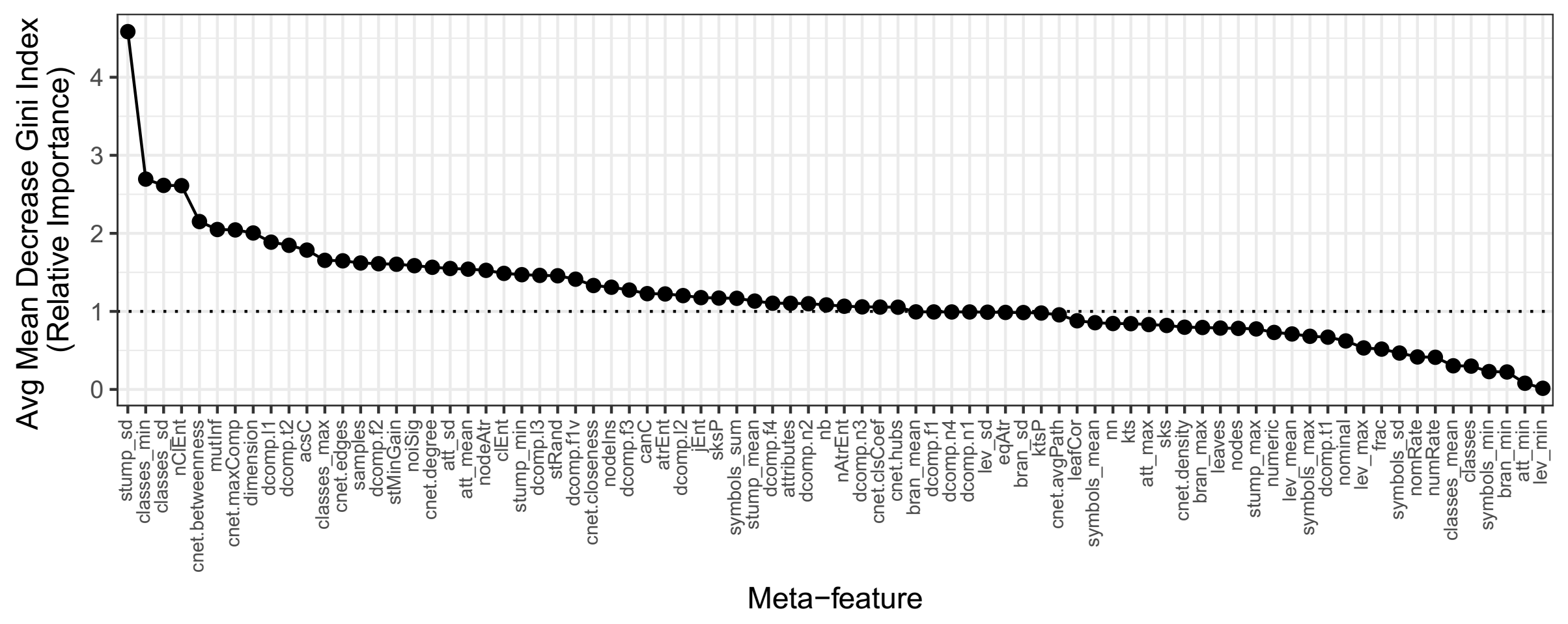}
        \label{fig:auc_gini}
    }
    \centering 
    \caption{Average meta-features relative importance obtained from \acrshort{rf} meta-models.~The names of the meta-features in the x-axis follow the acronyms presented in Tables~\ref{tab:mfeats_1} and \ref{tab:mfeats_2} in~\ref{app:features}.}
    \label{fig:rf-mfeat}
\end{figure}

From the induced \acrshort{rf} meta-models, the relative importance of the meta-features based on the Gini impurity index used to calculate the node splits~\cite{Breiman:2001}. Figure~\ref{fig:rf-mfeat} shows the average relative importance of the meta-features obtained from the \acrshort{rf} meta-models. The relative importance is shown for the experiments considering all meta-features and $\alpha=0.05$ (middle case).
At the x-axis, meta-features are presented in decreasing order according to their average relative importance values.
From this point, anytime a specific meta-feature is mentioned we present its name with a prefix indicating its category (according to Table~\ref{tab:mf-categories}).


Since no negative value (negative relative importance) was obtained, no meta-feature was discharged to build meta-models.
It also shows that a large number of meta-features were relevant to the induction of the meta-models, a possible reason for why meta-feature selection produced worse results for most of the meta-learners.

The most important meta-feature was a landmarking meta-feature: ``\texttt{LM.stump\_sd}'', which describes the standard deviation of the number of examples correctly classified by a decision stump. 
It measures the complexity of the problem considering its simplicity. 
The second most important was a simple meta-feature: ``\texttt{SM.classes\_min}'',  which measures the minority class size. The third was also a simple meta-feature: ``\texttt{SM.classes\_sd}'', which describes the standard deviation of the number of examples per class.
These meta-features together strongly indicate that for \acrshort{rf}, the most important meta-features are related with class imbalance. A rule extracted from a model induced by \acrshort{rf} states that if the dataset is imbalanced, it is better to use default \acrshort{hp} for \acrshortpl{svm}. The other important meta-features were:

\begin{itemize}
    
    \item ``\texttt{IN.nClEnt}'' and ``\texttt{IN.mutInf}'': these are information-theoretical meta-features. While the first describes the class entropy for a normalized base level dataset, the second measures the mutual information, a reduction of uncertainty about one random feature given the knowledge of another; 
    
    \item ``\texttt{CN.betweenness}'': betweenness centrality is a meta-feature derived from complex networks that measures, for a set of vertex and edges, the average number of shortest paths that traverses them. The value will be small for simple datasets, and high for complex datasets;
    
    \item ``\texttt{DC.l1}'' and ``\texttt{DC.t2}'': these are data complexity meta-features. While the first measures the minimum of an error function for a linear classifier, the second measures the average number of points per dimension. These features are related with the class separability (l1), and the geometry of the problem's dimension (t2);
    
    \item ``\texttt{SM.dimension}'': this is a simple meta-feature that measures the relation between the number of examples and attributes in a dataset;
    
    \item ``\texttt{CN.maxComp}'': this is another complex-network meta-feature. It measures the maximum number of connected components in a graph. If a dataset presents a high overlapping of its classes, the graph will present a large number of disconnected components, since connections between different classes are pruned.
    
\end{itemize}

\noindent Among the most important, there are meta-features from different categories (simple, data complexity, complex-networks and from information-theory). Complex-network measures describe data complexity regarding graphs and indicate how sparse the classes are between their levels. Data complexity meta-features try to extract information related to the class separability.
The stump meta-feature works along the same lines, trying to identify the complexity of the problem by simple landmarking. 
The information-theoretical meta-features indirectly checks how powerful the dataset attributes are to solve the classification problem.

Although summarized rules cannot be obtained from \acrshort{rf} meta-models, the analysis of meta-features importance provides some useful information. 
For instance, dataset characteristics such as the data balancing, class sizes, complexity and linearity were considered relevant to recommend when \acrshort{hp} tuning is required.


\subsection{Linearity Hypothesis}
\label{subsec:when}

The previous sections, in particular the \acrshort{rf} meta-analysis, suggest that linearity is a key aspect to decide between the recommendation of default or tuned \acrshortpl{hp} values. 
Experimental results indicate that default \acrshort{hp} values might be good for classification tasks with high linear separability. 
As a consequence, tuning would be required for tasks with complex decision surfaces, where \acrshortpl{svm} would need to find irregular decision boundaries.


\begin{figure*}[ht!]
    \subfloat[Performance differences between \acrshort{svm} and a linear classifier in all the base-level datasets.]
    {\includegraphics [width = \textwidth]
    {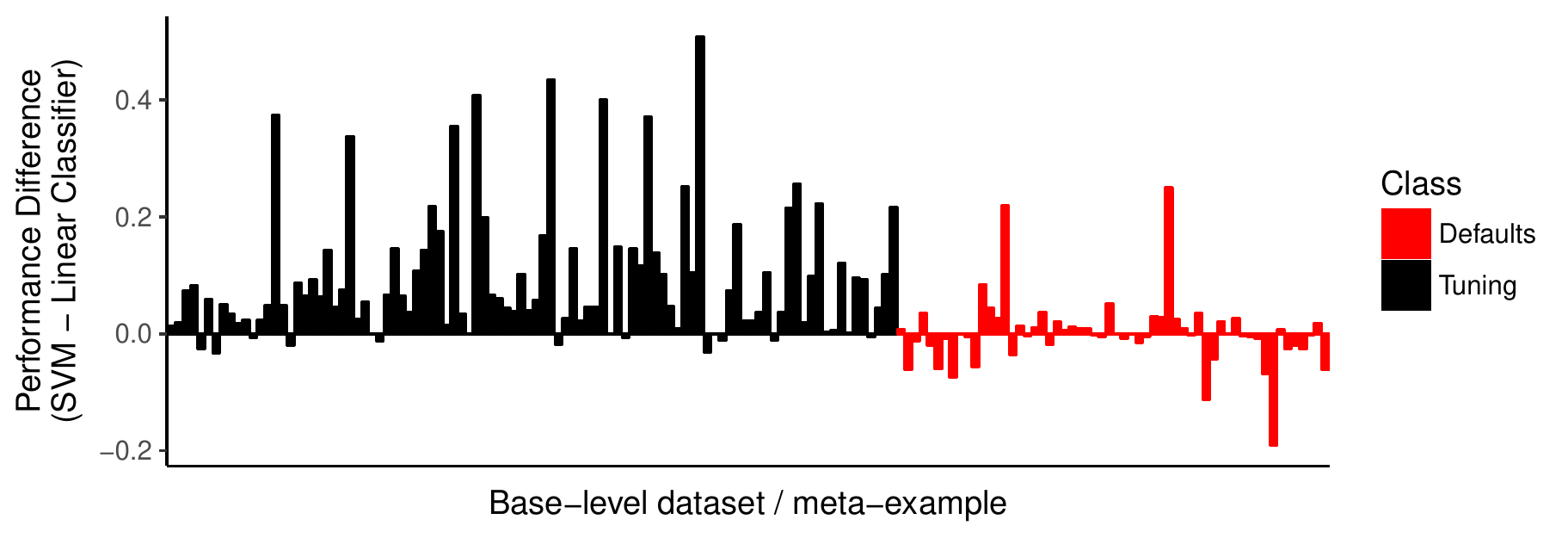}
       \label{fig:perf_diff}
    }
    \\
    \subfloat[Average relative importance of the meta-features obtained from \acrshort{rf} meta-models.~The names of the meta-features in the x-axis follow the acronyms presented in Tables~\ref{tab:mfeats_1} and \ref{tab:mfeats_2} in~\ref{app:features}.]
    {\includegraphics  [width = \textwidth]
    {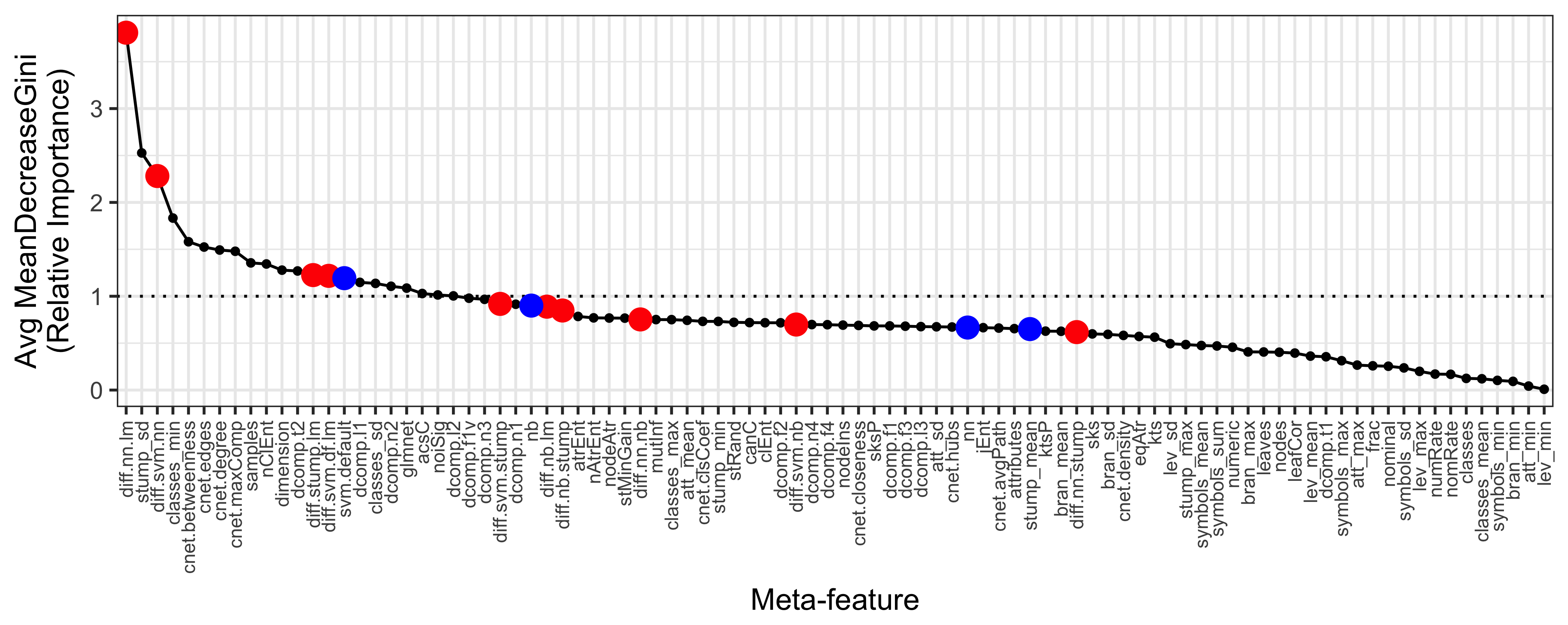}
        \label{fig:j48_avg_impr}
    }
    \caption{Linearity hypothesis results considering relative landmarking meta-features.}
    \label{fig:newFeat}
\end{figure*}


In order to investigate this hypothesis, a linear classifier was also evaluated in all the available $156$ datasets using the same base-level experimental setup described in Table~\ref{tab:hp_setup}. If the linearity hypothesis is true, the performance difference between \acrshortpl{svm} and the linear classifier in meta-examples labeled as ``\texttt{Defaults}'' would be smaller than or equal to the meta-examples labeled as ``\texttt{Tuning}'.

Figure~\ref{fig:perf_diff} shows the performance differences obtained in all the datasets at the base-level. 
Datasets at the x-axis are split based on their meta-target labels, ``\texttt{Tuning}', left side, in black, and ``\texttt{Defaults}'', right side, in red.
Despite some outliers, the performance differences for ``\texttt{Tuning}'' meta-examples are in general much higher than those for the ``\texttt{Defaults}'' meta-examples. Thus, the observed patterns support the linearity hypothesis.


In~\cite{Leite:2012}, the authors proposed a set of ``\acrfull{rl}'' meta-features based on the pairwise performance difference of simple landmarking algorithms. This new data characterizations schema is used to train meta-learning based on the \acrfull{at} algorithm.
The patterns observed in Figure~\ref{fig:perf_diff} follow the same principle, presenting a new alternative to characterize base-level datasets.
Following this proposal, $10$ new relative landmarking meta-features were generated based on five landmarking algorithms: \acrshort{knn}, \acrshort{nb}, \acrshort{lr}, \acrshort{svm} and \acrfull{ds}. These new meta-features are described in Table~\ref{tab:mfeats_2} in Appendix~\ref{app:features}.

The same \acrshort{rf} meta-analysis described in Section~\ref{subsec:rf} was performed, adding the relative landmarking meta-features to the meta-datasets. These experiments pointed out how useful the new meta-features are for the recommendation problem.
Figure~\ref{fig:newFeat} shows the relative importance values of the meta-features averaged in $30$ executions. 
The relative importance of these new meta-features are highlighted in red, while the simple landmarking is shown in blue. 

\begin{sloppypar}
Two of the relative landmarking meta-features are placed in the top-10 most important meta-features: \texttt{RL.diff.nn.lm} ($1ˆ{st}$), and \texttt{RL.diff.svm.lm} ($3ˆ{rd}$); another two measures are in the top-20 - \texttt{RL.diff.stump.lm} and \texttt{RL.diff.stump.lm}; and all of them depend directly on the linear classifier performance. It is also important to mention that simple landmarking meta-features performed, in general, worse than relative landmarking. All these relative importance plots show evidence that the linearity hypothesis is true, and at least one characteristic that defines the need of \acrshort{hp} tuning for \acrshortpl{svm} is the linearity of the base-level classification task.
\end{sloppypar}


\subsection{Overall comparison}

Given the potential shown by the relative landmarking meta-features, they were experimentally evaluated in combination with the 
meta-features previously evaluated as most important. 
Complex network (\texttt{cnet}) meta-features were included because they were ranked between the most important descriptors (as shown in Subsection~\ref{subsec:rf}). 
Simple and data complexity (\texttt{complex}) meta-features were the other two approaches evaluated in~the related studies listed in Section~\ref{subsec:tuning_necessity}.


\begin{figure}[ht!]
    \centering
    \subfloat[Average AUC values when considering relative landmarking meta-features with all the other meta-features' categories.]{
        \includegraphics
        [width=\textwidth]
        {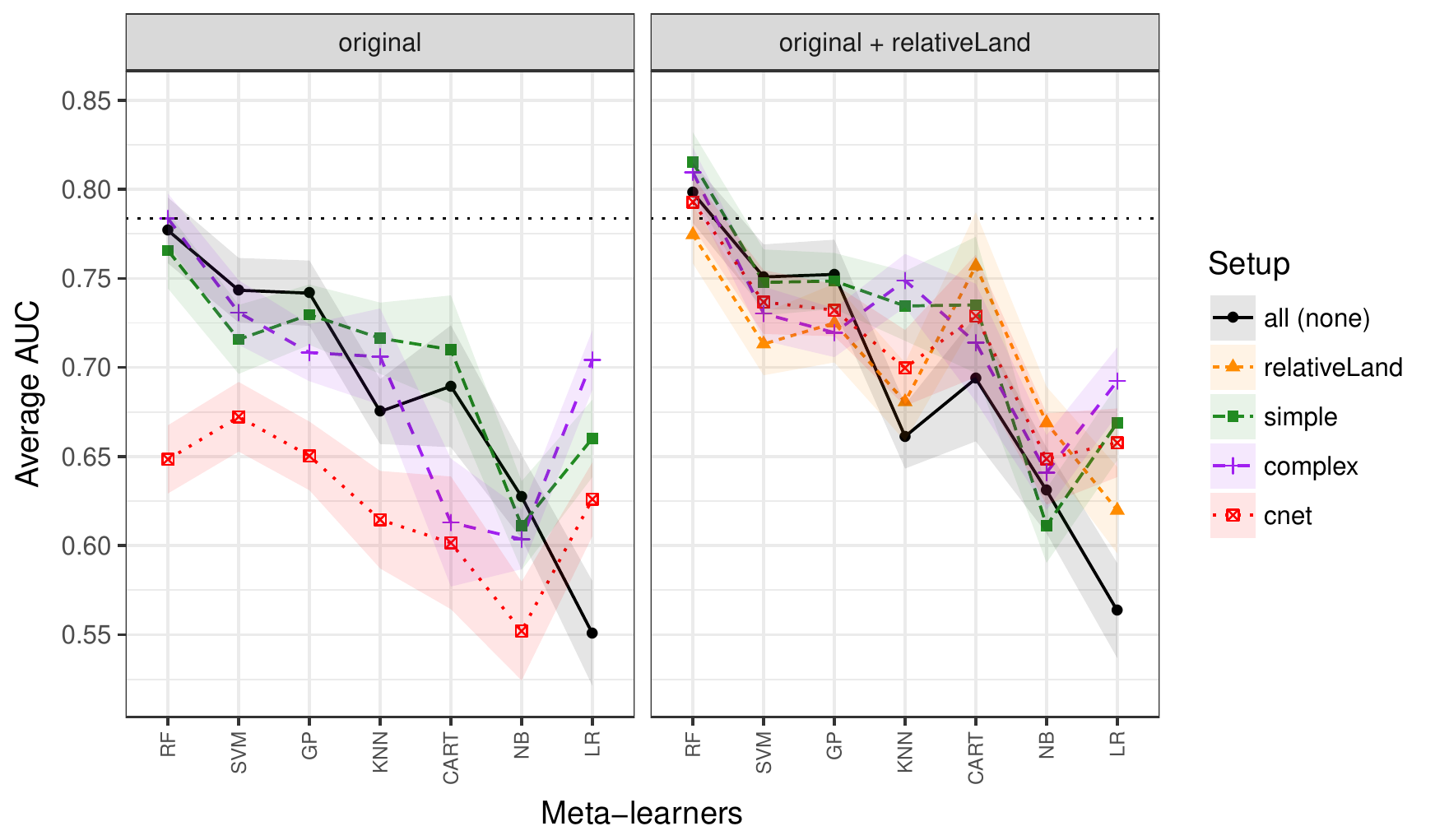}
        \label{fig:mfeat_relativeLand}
    }\\
    \subfloat[Average AUC values for the best overall experimental setup with the simple and the data complexity baselines presented in Section~\ref{sec:avg_perf}.]{
        \includegraphics
        [width = 0.66\textwidth]
        {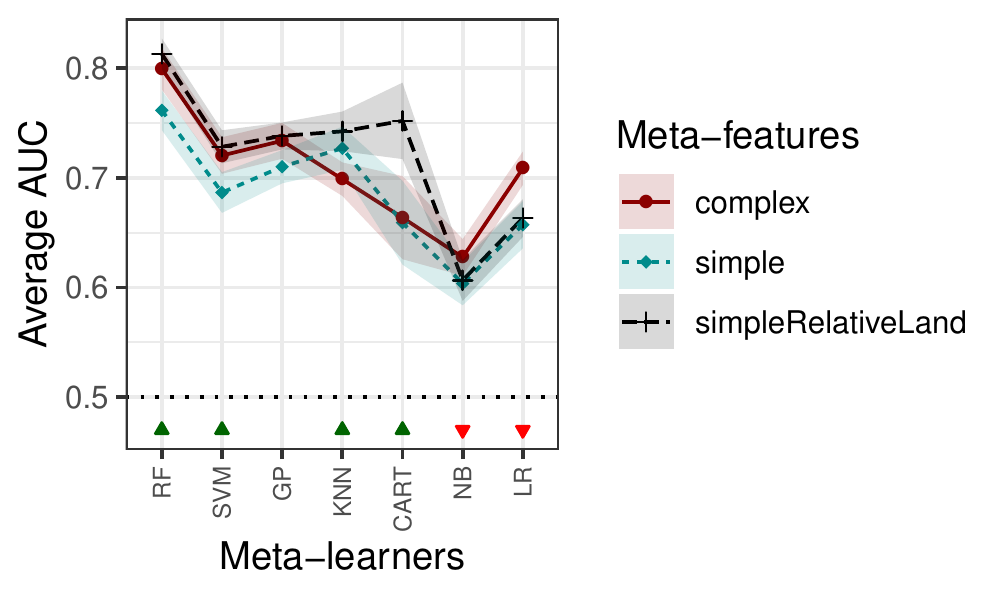}
        \label{fig:mfeat_significance}
    }
    \caption{Evaluating the previous experimental setups adding relative landmarking meta-features. The results are the average of $30$ runs.}
    \label{fig:comparison}
\end{figure}


Figure~\ref{fig:comparison} presents a comparison between the main experimental setups considering the addition of the relative landmarking (\texttt{relativeLand}) meta-features. The left chart of figure~\ref{fig:mfeat_relativeLand} shows \acrshort{auc} performance values obtained for each of the original setups, while the chart on the right presents setup performances when relative landmarking meta-features were included.
This figure shows that the use of relative landmarking meta-features improved all the setups where they were included.
At least three different setups used by \acrshort{rf} were higher than the \acrshort{auc} performance value obtained in the initial experiments. 
The setup considering simple and relative landmarking meta-features induced the best meta-models for \acrshort{rf}, \acrshort{svm} and \acrshortpl{gp}. The \acrshort{knn} and \acrshort{lr} meta-learners obtained the best predictive performance
using data complexity and relative landmarking meta-features and the same occurred for  \acrshort{cart} and \acrshort{nb} for ``relativeLand'' set.

\begin{sloppypar}

Figure~\ref{fig:mfeat_significance} compares the best setup from Figure~\ref{fig:mfeat_relativeLand}: ``\texttt{simpleRelativeLand}'', which uses both simple and relative landmarking meta-features, with the the baselines from Figure~\ref{fig:results}, using``\texttt{simple}'' and data complexity (``\texttt{complex}'') meta-features, often explored in related studies (see Table~\ref{tab:svm_related}).

In this figure, the x-axis shows the different meta-learners, while the y-axis shows their predictive performance assessed by the AUC averaged over $30$ repetitions. 
The Wilcoxon paired-test with $\alpha$ = 0.05 was applied to assess the statistical significance of these results. An upward green triangle (\textcolor{cadmiumgreen}{$\blacktriangle$}) at the x-axis identifies situations where the use of ``\texttt{simpleRelativeLand}'' was statistically better than using the baselines. In the same figure, the red downward triangles (\textcolor{cadmiumred}{$\blacktriangledown$}) indicate when the use of baselines was significantly better. In the remaining cases, the predictive performance of the meta-models were equivalents.
\end{sloppypar}

Overall, the meta-models induced with ``\texttt{simpleRelativeLand}'' meta-features were significantly better than those induced with baseline meta-features for most of the meta-learners: \acrshort{rf}, \acrshort{svm}, \acrshort{knn} and \acrshort{cart} obtained superior \acrshort{auc} values. Furthermore, the best meta-learner (RF) also significantly outperformed our previous results. 
The baselines produced the best meta-models for only two algorithms: \acrshort{nb} and \acrshort{lr}. For the \acrshort{gp} algorithm, the different setups did not present any statistically significant difference.


\subsection{Analysis of the predictions}

\begin{figure}[h!]
    \centering
    \subfloat[Meta-learners' individual predictions.]
    {
        \centering
        \includegraphics
        [width=0.9\textwidth]
        {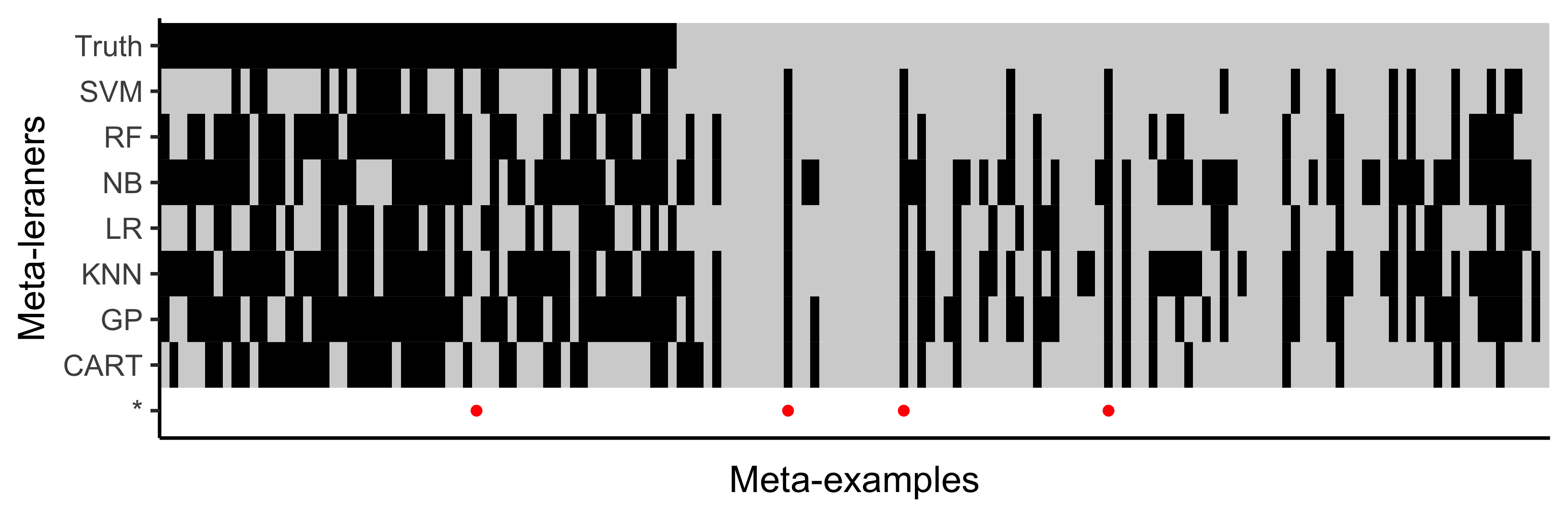}
        \label{fig:preds}
    }
    \\
    \subfloat[Meta-learners' misclassification rates.]
    {
        \hspace*{1.4cm}
        \includegraphics
        [scale = 0.8]
        {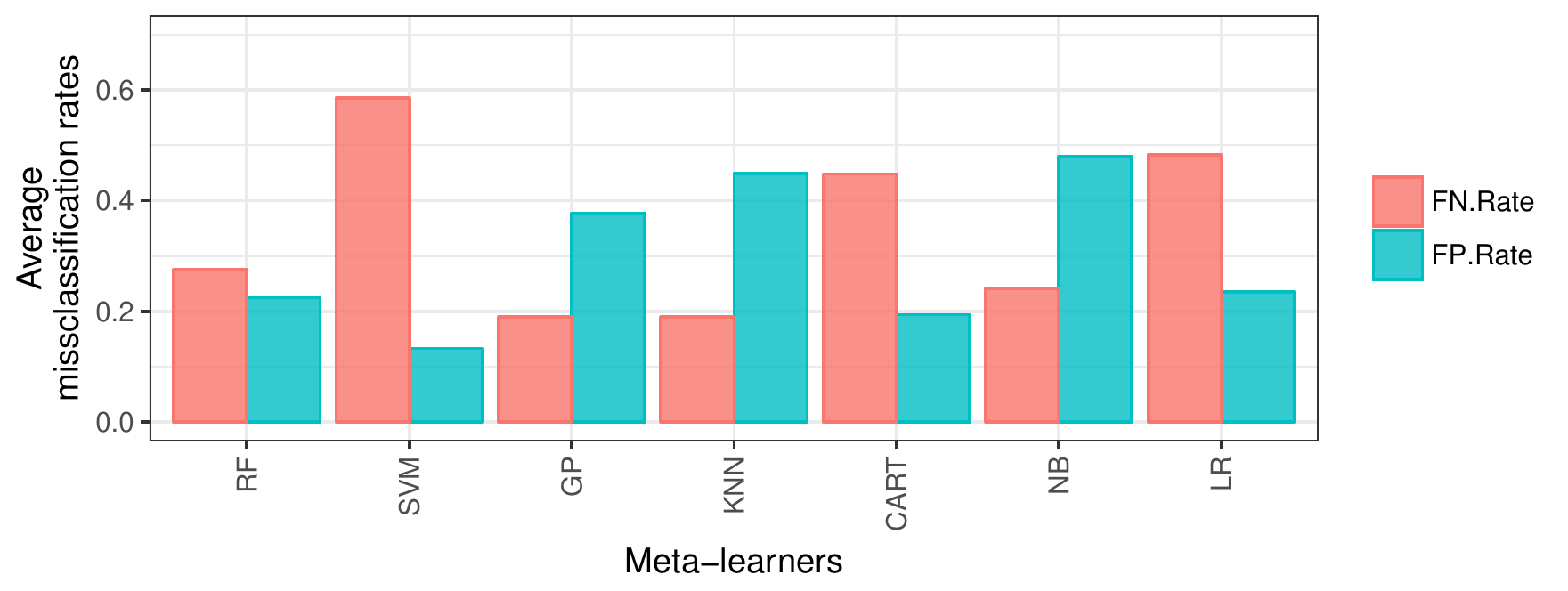}
        \label{fig:err}
    }
    \caption{Meta-learners' predictions considering the experimental setups which obtained the best \acrshort{auc} values.}
    \label{fig:pred_all}
\end{figure}

A more in-depth analysis of the meta-learner predictions can help to understand their behavior.
Figure~\ref{fig:pred_all} shows the misclassifications of the meta-learners considering their best experimental setups. 
The top chart (Fig. ~\ref{fig:preds}) shows all the individual predictions, with the x-axis listing all the meta-examples and y-axis the meta-learners. 
In this figure, ``\texttt{Defaults}'' labels are shown in black and ``\texttt{Tuning}'' labels in gray.
The top line in the y-axis, ``\texttt{Truth}'' shows the truth labels of the meta-examples, which are ordered according to their truth labels. 
The bottom line (``*'') shows red points for meta-examples misclassified by all meta-learners. 


In the \acrshort{svm} \acrshort{hp} tuning recommendation task, ``Defaults'' is defined as the positive class and ``\texttt{Tuning}'' as the negative class. Therefore,  a \acrshort{fn} is a wrong recommendation to perform \acrshort{hp} tuning on \acrshortpl{svm}, and \acrfull{fp} is a wrong recommendation to use default \acrshort{hp} values. While a reduction in \acrshort{fn} can decrease the computational cost, a reduction in
\acrshort{fp} can improve predictive performance. 

Algorithms following different learning biases present different prediction patterns and this can be observed in Figure ~\ref{fig:pred_all}. 
Usually, most meta-examples are correctly classified (a better performance than the baselines).
Besides, the following patterns can be observed:

\begin{itemize}
    
    \item \acrshort{knn} and \acrshort{gp} minimize the \acrshort{fn} rate, correctly classifying most of the meta-examples as ``\texttt{Defaults}''.
    However, they misclassified many examples from the ``\texttt{Tuning}'' class, penalizing the overall performance of the recommender system; 
    
    \item \acrshort{svm}, \acrshort{cart} and \acrshort{lr} minimized the \acrshort{fp} rate, correctly classifying most of the meta-examples requiring tuning. However, they tended to classify the meta-examples in the majority class;

\end{itemize}

\noindent A more balanced scenario is provided by the \acrshort{rf} meta-models, which presented the best predictive performance. Although it was not the best algorithm for each class individually, it was the best when the two classes were considered.


\begin{table}[th!]

\setlength{\tabcolsep}{2pt}
\scriptsize
\centering
 \caption{Misclassified datasets by all the meta-learners. For each dataset it is shown: the meta-example number (Nro); the OpenML dataset name (Name) and id (id); the number of attributes (D), examples (N) and classes (C); the proportion between the number of examples from minority and majority classes (P); the performance values obtained by defaults (Def) and tuned (Tun) \acrshort{hp} settings assessed by \acrshort{bac}; and the truth label (Label).}

\begin{tabular}{clrrrrcccc}
  \toprule
  
 \textbf{Nro } & \textbf{ Name } & \textbf{ id } & \textbf{ D } & \textbf{ N } & \textbf{ C } & \textbf{ P } & \textbf{ Def (sd) } & \textbf{ Tun (sd) } & \textbf{ Label } \\

    \midrule
    \rule{0pt}{2ex}
    
    17 & jEdit\_4.0\_4.2 & 1073 & 8 & 274 & 2 & 0.96 & 0.73 \footnotesize{(0.01)} & 0.73 \footnotesize{(0.01)} & \texttt{Defaults} \\

    36 & banknote-authentication & 1462 & 4 & 1372 & 2 & 0.80 & 0.99 \footnotesize{(0.01)} & 0.99 \footnotesize{(0.01)} & \texttt{Tuning} \\

    78 & autoUniv-au7-500 & 1554 & 12 & 500 & 5 & 0.22 & 0.29 \footnotesize{(0.01)} & 0.31 \footnotesize{(0.01)} & \texttt{Tuning} \\
    
    97 & optdigits & 28 & 62 & 5620 & 10 & 0.97 & 0.99 \footnotesize{(0.01)} & 0.99 \footnotesize{(0.01)} & \texttt{Tuning} \\
    
    \bottomrule
\end{tabular}
\label{tab:misclassified}
\end{table}

Table~\ref{tab:misclassified} lists all the datasets misclassified by all the meta-learners as indicated in Figure~\ref{fig:preds}. Meta-examples with ids $17$ (``\texttt{Defaults}'') and $78$ (``\texttt{Tuning}'') were corrected labeled by the statistical meta-rule, and therefore the misclassification may have occurred due to the lack of the descriptive ability of meta-features or noise in the meta-dataset. 
The other two meta-examples ($36$, $97$) were both labeled as ``\texttt{Tuning}'' but the statistical difference indicated is very small in terms of performance, and may indicate a limitation of the current meta-target rule criteria.


\subsection{Projecting performances at base-level}

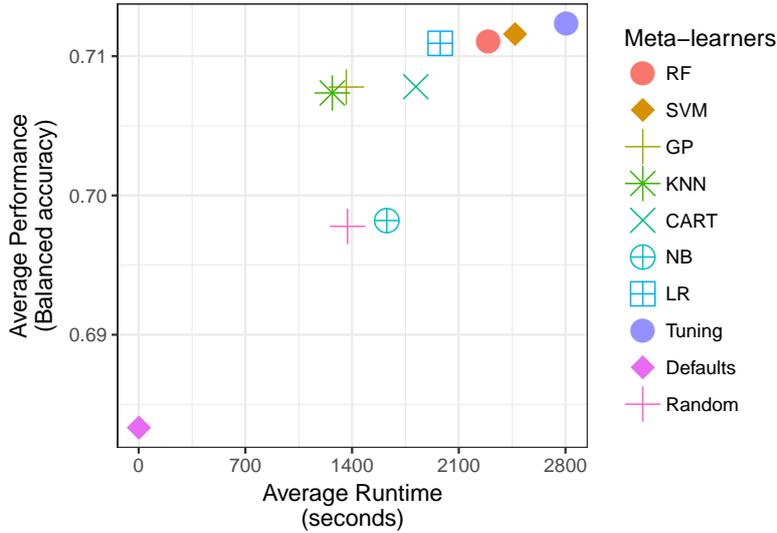
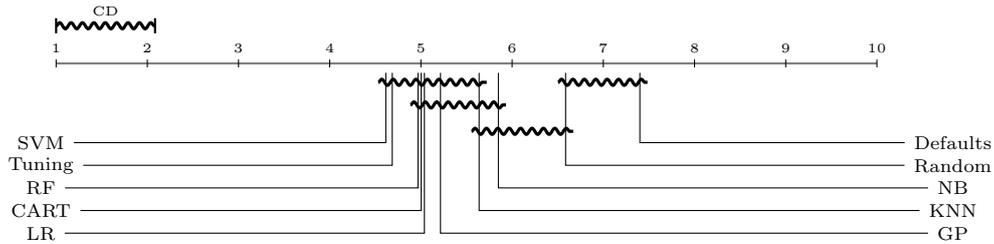
\begin{figure}[h!]
\centering 
    
    \subfloat[Average \acrshort{bac} and runtime 
    for the \acrshort{svm} base-level data.]
    {
    \includegraphics
    [scale = 0.8]
    {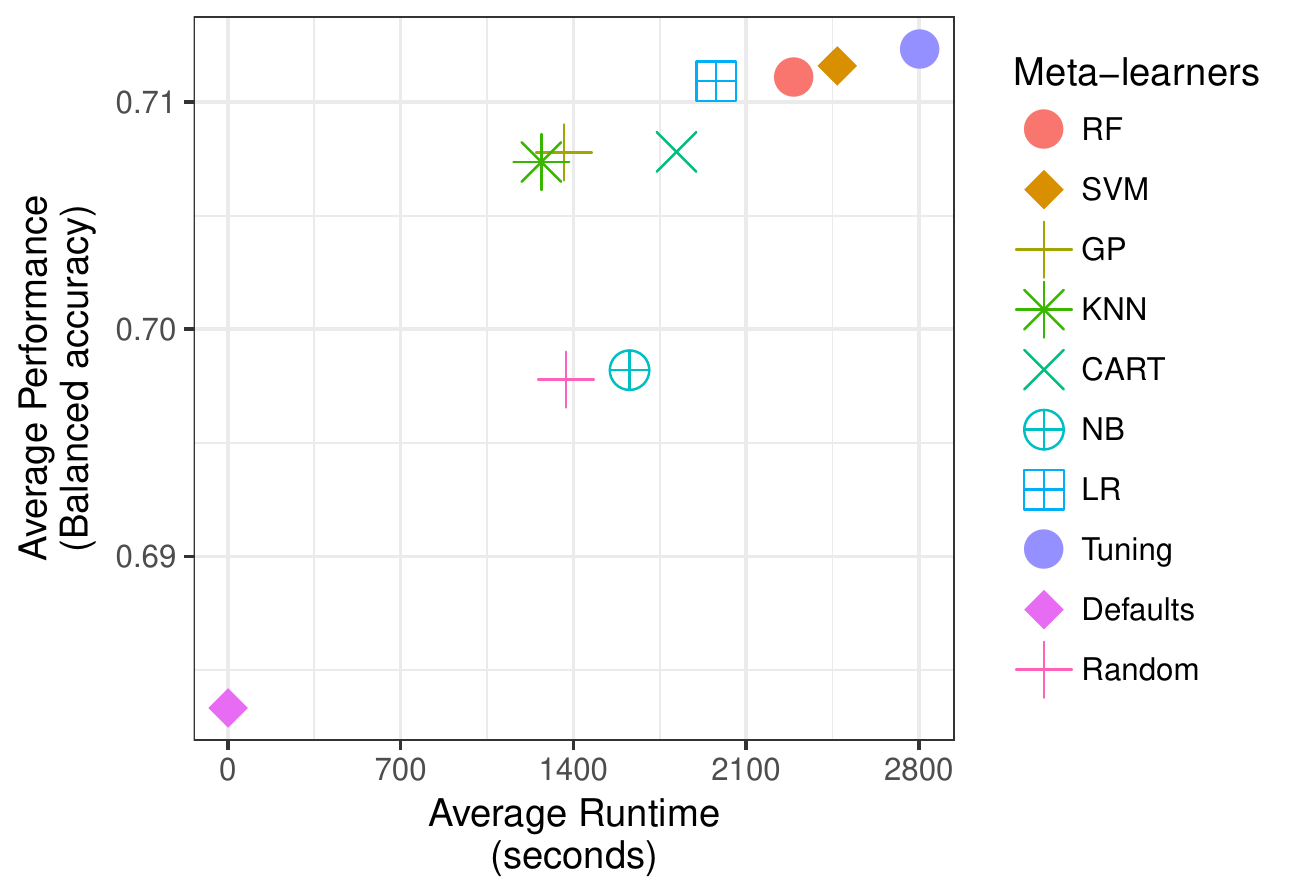}%
        \label{fig:perf_projec}
    } 
    \\
    \subfloat[\acrshort{cd} diagram comparing the \acrshort{bac} values of the meta-learners at the base-level according to the Friedman-Nemenyi test ($\alpha = 0.05)$.] 
    {
       \begin{tikzpicture}[xscale=2]
        \node (Label) at  (00.9254,0.7) {\tiny{CD}}; 
        \draw[decorate,decoration={snake,amplitude=.4mm,segment length=1.5mm,post length=0mm}, very thick, color = black](00.6000, 0.5) -- (01.2508, 0.5);
        \foreach \x in {00.6000,01.2508} \draw[thick,color = black] (\x, 0.4) -- (\x, 0.6);
        
        \draw[gray, thick](00.6000, 0) -- (06.0000, 0);
        \foreach \x in {00.6000,01.2000,01.8000,02.4000,03.0000,03.6000,04.2000,04.8000,05.4000,06.0000}\draw (\x cm,1.5pt) -- (\x cm, -1.5pt);
        \node (Label) at (00.6000,0.2) {\tiny{1}};
        \node (Label) at (01.2000,0.2) {\tiny{2}};
        \node (Label) at (01.8000,0.2) {\tiny{3}};
        \node (Label) at (02.4000,0.2) {\tiny{4}};
        \node (Label) at (03.0000,0.2) {\tiny{5}};
        \node (Label) at (03.6000,0.2) {\tiny{6}};
        \node (Label) at (04.2000,0.2) {\tiny{7}};
        \node (Label) at (04.8000,0.2) {\tiny{8}};
        \node (Label) at (05.4000,0.2) {\tiny{9}};
        \node (Label) at (06.0000,0.2) {\tiny{10}};
        \draw[decorate,decoration={snake,amplitude=.4mm,segment length=1.5mm,post length=0mm}, very thick, color = black](02.7192,-00.2500) -- ( 03.4327,-00.2500);
        \draw[decorate,decoration={snake,amplitude=.4mm,segment length=1.5mm,post length=0mm}, very thick, color = black](02.9307,-00.5500) -- ( 03.5596,-00.5500);
        \draw[decorate,decoration={snake,amplitude=.4mm,segment length=1.5mm,post length=0mm}, very thick, color = black](03.3327,-00.9000) -- ( 04.0019,-00.9000);
        \draw[decorate,decoration={snake,amplitude=.4mm,segment length=1.5mm,post length=0mm}, very thick, color = black](03.9019,-00.2500) -- ( 04.4904,-00.2500);
        \node (Point) at (02.7692, 0){};  \node (Label) at (0.5,-01.0500){\scriptsize{SVM}}; \draw (Point) |- (Label);
        \node (Point) at (02.8115, 0){};  \node (Label) at (0.5,-01.3500){\scriptsize{Tuning}}; \draw (Point) |- (Label);
        \node (Point) at (02.9807, 0){};  \node (Label) at (0.5,-01.6500){\scriptsize{RF}}; \draw (Point) |- (Label);
        \node (Point) at (03.0019, 0){};  \node (Label) at (0.5,-01.9500){\scriptsize{CART}}; \draw (Point) |- (Label);
        \node (Point) at (03.0230, 0){};  \node (Label) at (0.5,-02.2500){\scriptsize{LR}}; \draw (Point) |- (Label);
        \node (Point) at (04.4404, 0){};  \node (Label) at (6.5,-01.0500){\scriptsize{Defaults}}; \draw (Point) |- (Label);
        \node (Point) at (03.9519, 0){};  \node (Label) at (6.5,-01.3500){\scriptsize{Random}}; \draw (Point) |- (Label);
        \node (Point) at (03.5096, 0){};  \node (Label) at (6.5,-01.6500){\scriptsize{NB}}; \draw (Point) |- (Label);
        \node (Point) at (03.3827, 0){};  \node (Label) at (6.5,-01.9500){\scriptsize{KNN}}; \draw (Point) |- (Label);
        \node (Point) at (03.1288, 0){};  \node (Label) at (6.5,-02.2500){\scriptsize{GP}}; \draw (Point) |- (Label);
        \end{tikzpicture}

       \label{fig:friedman_test}
    }
    \caption{Performance of the meta-learners projected into the \acrshort{svm} hyperparameter tuning problem (base-level).}
    \label{fig:final}
\end{figure}

This section assesses the impact of the choices made by the meta-learners at the base-level. It also analyses and discusses the reduction of runtime when using the proposed meta-learning recommender system. 
Figure~\ref{fig:final} shows the predictive performance of \acrshortpl{svm} at the base-level using the method (``\texttt{Tuning}' or ``\texttt{Defaults}') selected by the meta-learners to define their \acrshort{hp} values. The best meta-learners identified in the previous sections were compared with three simple baselines: a model that always recommends \acrshort{hp} tuning (\texttt{Tuning}), a model that always recommends the use of defaults (\texttt{Defaults}), and a model that provides random recommendations (\texttt{Random}).

Figure~\ref{fig:perf_projec} shows a scatter plot with the projected \acrshort{bac} performance and runtime value averaged in all the base-level datasets. Performing always the \acrshort{hp} tuning had the highest average \acrshort{bac} value but was also the most expensive approach. On the other hand, always using default \acrshort{hp} values is the fastest approach, but with the lowest average  \acrshort{bac} value.
The proposed meta-models are above \texttt{Random} and \texttt{Defaults} baselines, performing close to the \texttt{Tuning} baseline but with lower average runtime costs.

The Friedman test, with a significance level of $\alpha = 0.05$, was also used to assess the statistical significance of the base-level predictions. The null hypothesis is that all the meta-learners and baselines are equivalent regarding the average predictive \acrshort{bac} performance. When the null hypothesis is rejected, the Nemenyi post-hoc test is  applied to indicate when two different techniques are significantly different. Figure~\ref{fig:friedman_test} presents the \acrfull{cd} diagram. Techniques are connected when there is \textit{no} significant differences between them. 

Overall, the approach always using default \acrshortpl{hp} (\texttt{Defaults}) is ranked last, followed by the \texttt{Random} baseline. Almost all meta-learners are significantly better than both and are equivalent to \texttt{Tuning}, which always performs \acrshort{hp} tuning.
In this figure, although the \acrshort{rf} meta-model is considered the best,
it was not ranked first. The first was the \acrshort{svm} meta-model. This occurred because it most often recommended the use of tuned settings, which was reflected in its performance at the base-level.
With many datasets at the base-level, it can be pointed out that
the overall gain is diluted between them. Even so, the meta-learners could considerably reduce the computational costs related to tuning, maintaining a high predictive performance.


Besides, it can be observed in Figure~\ref{fig:perf_projec} that tuning \acrshort{svm} hyperparameters for just one dataset will take on average two days. The most costly datasets (with a high number of features or examples) took almost 10 days to finish all the 10 tuning repetitions (seeds) even paralleling the jobs in a high performance cluster. Regarding the meta-feature extraction, the same datasets will take at most 10 minutes, specially because of some mathematical operations used by~\acrfull{dc}~meta-features. For most meta-features, the time taken to characterize a dataset is less than $30$ seconds. Thus, during the prediction phase with the induced meta-model, the computational cost of extracting the characteristics of a new dataset is irrelevant compared to the computational cost of the tuning process. We think this is an important argument in favor of using our system, which is used in practical scenarios. 


\subsection{A note on the generalization}
\label{sec:generalization}

Although the main focus of the paper is to investigate the \acrshort{svm} hyperparameter tuning problem, we also conducted experiments for predicting the need for tuning~\acrfullpl{dt}. 
These experiments aim to provide more evidence that the proposed method can be generalized. 
Once the meta-knowledge is extracted, the system is able to induce meta-models to any supervised learning algorithm.
In particular, we investigated the \acrshort{hp} tuning of the \texttt{J48} algorithm, a \texttt{WEKA}\footnote{\url{http://weka.sourceforge.net/doc.dev/weka/classifiers/trees/J48.html}} implementation for the Quinlan`s C4.5~\acrshort{dt} induction algorithm~\cite{Quinlan:1993}, and one of the most popular \acrshort{ml} algorithms. 
The tree models were induced using the \texttt{RWeka}\footnote{\url{https://cran.r-project.org/web/packages/RWeka/index.html} package.

The meta-datasets for the J48 algorithm were generated based on \acrshort{hp} tuning results obtained from $102$ datasets reported in~\cite{Mantovani:2016}}.~We also expanded the meta-knowledge by performing additional experiments to cover the same datasets explored with~\acrshortpl{svm}. In total, we obtained results performing the \acrshort{hp} tuning of the J48 algorithm in $165$ \acrshort{openml} datasets. The tuning processes followed the experimental setup described in Table~\ref{tab:hp_setup} with some differences:
\begin{itemize}
    \item the J48 hyperparameter space has nine hyperparameters\footnote{The J48 hyperparameter space is also presented in~\ref{app:j48_space}.};
    \item a budget size of $900$ evaluations was adopted in the experiments with trees. It is greater than for \acrshortpl{svm} because of the greater search space of J48;
    \item the tuned hyperparameter results were compared with those obtained from the \texttt{J48} \texttt{RWeka}/\texttt{WEKA} default settings.
\end{itemize}


Table~\ref{tab:j48_datasets} in~\ref{app:j48_space} presents the main characteristics of the meta-datasets generated with J48 tuning experiments. Class distribution columns (\textit{Tuning} and \textit{Default}) show values which indicate a different \textit{hyperparameter profile}\footnote{In this paper, the ``\textit{hyperparameter profile}'' term refers to how sensitive an algorithm may be to the \acrshort{hp} tuning task.} compared to that observed in experiments with~\acrshortpl{svm}: most of the meta-examples are labelled as ``\texttt{Default}'', i.e., tuning did not statistically improve the algorithm performance in two thirds of the datasets. 
Here, we present results just on the meta-dataset using $\alpha$ = 0.05 for the statistical labeling rule. However, results obtained with different $\alpha$ values were quite similar.


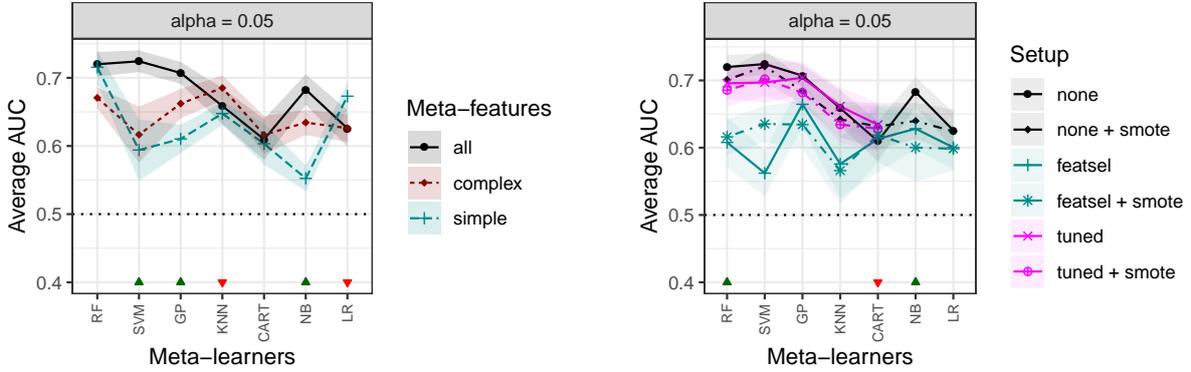
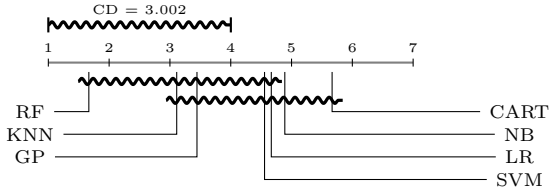
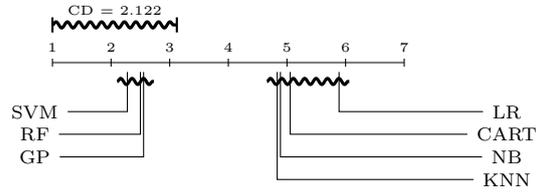
\begin{figure}[h!]
    \centering
    \subfloat[AUC values considering different categories of meta-features.]
    {
        \centering
        \includegraphics
        [width=0.46\textwidth]
        {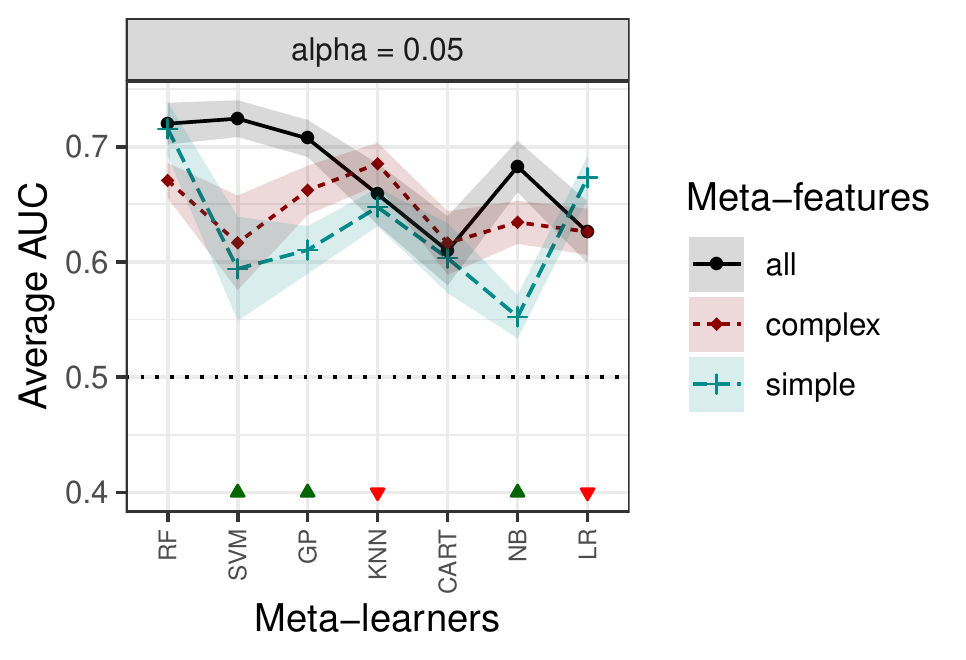}
        \label{fig:j48_mtl_auc}
    }
    \hspace{0.4cm}
    \subfloat[AUC values considering different experimental setups.]
    {
        \includegraphics
        [width=0.46\textwidth]
        {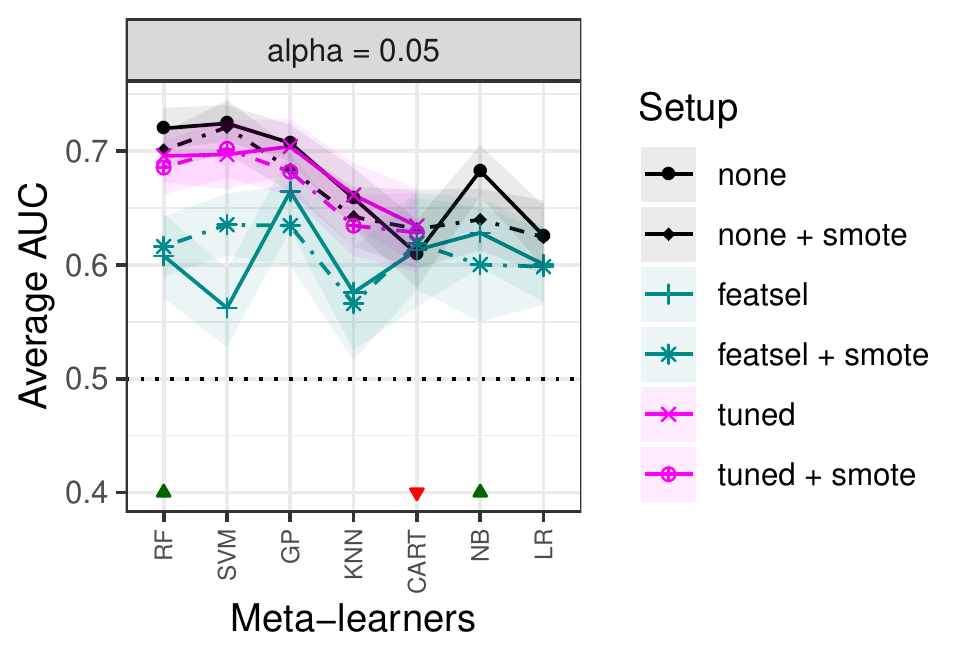}
        \label{fig:j48_mtl_setups}
    }
    \\
    \subfloat[CD diagram comparing different meta-learners in different categories of meta-features.]
    {
        \begin{tikzpicture}[xscale=2.8]

            \node (Label) at  (00.7146,0.7) {\tiny{CD = 3.002}}; 
            \draw[decorate,decoration={snake,amplitude=.4mm,segment length=1.5mm,post length=0mm}, very thick, color = black](00.2857, 0.5) -- (01.1434, 0.5);
            \foreach \x in {00.2857,01.1434} \draw[thick,color = black] (\x, 0.4) -- (\x, 0.6);
            
            \draw[gray, thick](00.2857, 0) -- (02.0000, 0);
            \foreach \x in {00.2857,00.5714,00.8571,01.1429,01.4286,01.7143,02.0000}\draw (\x cm,1.5pt) -- (\x cm, -1.5pt);
            \node (Label) at (00.2857,0.2) {\tiny{1}};
            \node (Label) at (00.5714,0.2) {\tiny{2}};
            \node (Label) at (00.8571,0.2) {\tiny{3}};
            \node (Label) at (01.1429,0.2) {\tiny{4}};
            \node (Label) at (01.4286,0.2) {\tiny{5}};
            \node (Label) at (01.7143,0.2) {\tiny{6}};
            \node (Label) at (02.0000,0.2) {\tiny{7}};
            \draw[decorate,decoration={snake,amplitude=.4mm,segment length=1.5mm,post length=0mm}, very thick, color = black](00.4260,-00.2500) -- ( 01.3833,-00.2500);
            \draw[decorate,decoration={snake,amplitude=.4mm,segment length=1.5mm,post length=0mm}, very thick, color = black](00.8389,-00.5000) -- ( 01.6690,-00.5000);
            \node (Point) at (00.4760, 0){};  \node (Label) at (0.2,-00.6500){\scriptsize{RF}}; \draw (Point) |- (Label);
            \node (Point) at (00.8889, 0){};  \node (Label) at (0.2,-00.9500){\scriptsize{KNN}}; \draw (Point) |- (Label);
            \node (Point) at (00.9841, 0){};  \node (Label) at (0.2,-01.2500){\scriptsize{GP}}; \draw (Point) |- (Label);
            \node (Point) at (01.6190, 0){};  \node (Label) at (2.5,-00.6500){\scriptsize{CART}}; \draw (Point) |- (Label);
            \node (Point) at (01.3968, 0){};  \node (Label) at (2.5,-00.9500){\scriptsize{NB}}; \draw (Point) |- (Label);
            \node (Point) at (01.3333, 0){};  \node (Label) at (2.5,-01.2500){\scriptsize{LR}}; \draw (Point) |- (Label);
            \node (Point) at (01.3016, 0){};  \node (Label) at (2.5,-01.5500){\scriptsize{SVM}}; \draw (Point) |- (Label);
        
        \end{tikzpicture}
        \label{fig:j48_friedman_mfeats}
    }
    \hspace{0.4cm}
    \subfloat[CD diagram comparing different meta-learners in different experimental setups).]
    {

        \begin{tikzpicture}[xscale=2.7]

               \node (Label) at  (00.5889,0.7) {\tiny{CD = 2.122}}; 
            \draw[decorate,decoration={snake,amplitude=.4mm,segment length=1.5mm,post length=0mm}, very thick, color = black](00.2857, 0.5) -- (00.8920, 0.5);
            \foreach \x in {00.2857,00.8920} \draw[thick,color = black] (\x, 0.4) -- (\x, 0.6);
            
            \draw[gray, thick](00.2857, 0) -- (02.0000, 0);
            \foreach \x in {00.2857,00.5714,00.8571,01.1429,01.4286,01.7143,02.0000}\draw (\x cm,1.5pt) -- (\x cm, -1.5pt);
            \node (Label) at (00.2857,0.2) {\tiny{1}};
            \node (Label) at (00.5714,0.2) {\tiny{2}};
            \node (Label) at (00.8571,0.2) {\tiny{3}};
            \node (Label) at (01.1429,0.2) {\tiny{4}};
            \node (Label) at (01.4286,0.2) {\tiny{5}};
            \node (Label) at (01.7143,0.2) {\tiny{6}};
            \node (Label) at (02.0000,0.2) {\tiny{7}};
            \draw[decorate,decoration={snake,amplitude=.4mm,segment length=1.5mm,post length=0mm}, very thick, color = black](00.6008,-00.2500) -- ( 00.7800,-00.2500);
            \draw[decorate,decoration={snake,amplitude=.4mm,segment length=1.5mm,post length=0mm}, very thick, color = black](01.3309,-00.2500) -- ( 01.7323,-00.2500);
            \node (Point) at (00.6508, 0){};  \node (Label) at (0.2,-00.6500){\scriptsize{SVM}}; \draw (Point) |- (Label);
            \node (Point) at (00.7143, 0){};  \node (Label) at (0.2,-00.9500){\scriptsize{RF}}; \draw (Point) |- (Label);
            \node (Point) at (00.7300, 0){};  \node (Label) at (0.2,-01.2500){\scriptsize{GP}}; \draw (Point) |- (Label);
            \node (Point) at (01.6823, 0){};  \node (Label) at (2.5,-00.6500){\scriptsize{LR}}; \draw (Point) |- (Label);
            \node (Point) at (01.4443, 0){};  \node (Label) at (2.5,-00.9500){\scriptsize{CART}}; \draw (Point) |- (Label);
            \node (Point) at (01.3968, 0){};  \node (Label) at (2.5,-01.2500){\scriptsize{NB}}; \draw (Point) |- (Label);
            \node (Point) at (01.3809, 0){};  \node (Label) at (2.5,-01.5500){\scriptsize{KNN}}; \draw (Point) |- (Label);
            
            \end{tikzpicture}

        \label{fig:j48_friedman_setups}
    }
    \caption{Meta-learners average AUC results in the J48 meta-dataset labelled with $alpha=0.05$, and CD diagrams comparing meta-learners according to the Friedman-Nemenyi test ($\alpha=0.05$). Results are averaged in $30$ runs.}
    \label{fig:mtl_dt_results}
\end{figure}



The predictive performance of the meta-learners considering different categories of meta-features are summarized in Figure~\ref{fig:j48_mtl_auc}. The best results were obtained using the \acrshort{rf}, \acrshort{svm} and \acrshort{gp}  meta-learners. 
They achieved their best predictive performances using preferably ``all'' the available meta-features, with \acrshort{auc} values between \acrshort{auc}$\:=(0.67,0.72)$. It may be due to the fact that predictions tend toward a specific class (Defaults) since these meta-datasets are imbalanced.
Overall, the best meta-model was obtained by the \acrshort{svm} algorithm. However, when considering all the possible scenarios (different $\alpha$ values), there was no statistical difference among the top ranked meta-learners: the \acrshort{rf}, \acrshort{knn}, \acrshort{gp}, \acrshort{svm} and \acrshort{lr} algorithms (see Figure~\ref{fig:j48_friedman_mfeats}). Furthermore, in general, the complete set of meta-features provided the best results for most algorithms. 

Figure~\ref{fig:j48_mtl_setups} shows the average \acrshort{auc} values considering different experimental setups. The \texttt{NB} and \texttt{LR} algorithms do not have any tunable hyperparameters. Consequently, their results for ``\texttt{tuned}'' setups are missing in the chart.
In general, when considering \acrshort{auc} values obtained in the original meta-dataset, 
improvements were obtained only for \acrshort{cart} and \acrshort{knn} applying \acrshort{smote}. For the other algorithms, best meta-learners were still induced without any additional process. Statistical comparisons highlight the performance of the \acrshort{svm}, \acrshort{rf} and \acrshort{gp} algorithms (see Figure~\ref{fig:j48_friedman_setups}). 

We also evaluated the potentiality of the \acrfull{rl} meta-features in the J48 tuning recommendation problem. However, differently than reported in Section~\ref{subsec:when}, they worsened most of the meta-models. Linearity is not a key aspect in the J48 tuning problem, which further reinforces the results obtained from \acrshortpl{svm} (see Section~\ref{subsec:when}). 

Reproducing the \acrshort{rf} analysis in the J48 tuning problem (see Section~\ref{subsec:rf}), the most important meta-feature was the data complexity measure ``\texttt{DC.f4}''. This meta-feature describes the collective attribute efficiency in a dataset. The second was a simple meta-feature: ``\texttt{SM.abs\_cor}'', a metric that measures the linear relationship between two attributes. This value is averaged in all pairs of attributes in the dataset. The top-3 is completed with another data complexity meta-feature: ``\texttt{DC.f3}'', which describes the maximum individual attribute efficiency. The two \acrshort{dc} meta-features measure the discriminative power of the dataset's attributes, while the absolute correlation verifies if the information provided by attributes is not redundant. These most important meta-features suggest that if a dataset has representative attributes, default \acrshort{hp} values are robust to solve it. Otherwise, the J48 tuning is recommended. 


\begin{figure}[h!]
    \centering
    \subfloat[Average \acrshort{bac} and runtime 
    for the J48 base-level data.]
    {
        \centering
        \includegraphics
        [scale = 0.7]
        {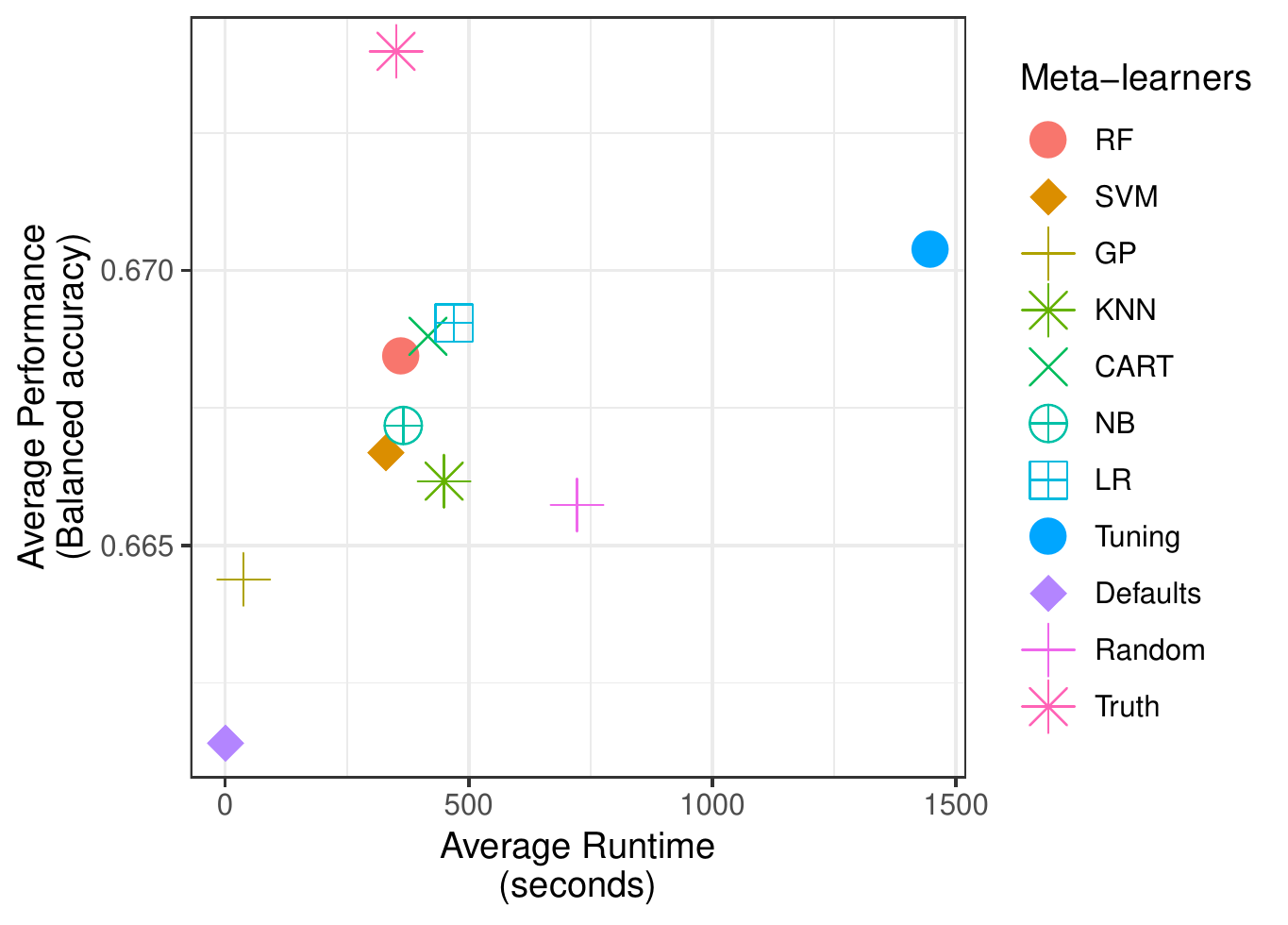}
        \label{fig:j48_projections}
    }
    \\
    \subfloat[\acrshort{cd} diagram comparing the \acrshort{bac} values of the meta-learners at the base-level according to the Friedman-Nemenyi test ($\alpha = 0.05)$.] 
    {
      \begin{tikzpicture}[xscale=5.6]

            \node (Label) at  (00.3055,0.7) {\tiny{CD = 1.0546}}; 
            \draw[decorate,decoration={snake,amplitude=.4mm,segment length=1.5mm,post length=0mm}, very thick, color = black](00.2000, 0.5) -- (00.4109, 0.5);
            \foreach \x in {00.2000,00.4109} \draw[thick,color = black] (\x, 0.4) -- (\x, 0.6);
            
            \draw[gray, thick](00.2000, 0) -- (02.0000, 0);
            \foreach \x in {00.2000,00.4000,00.6000,00.8000,01.0000,01.2000,01.4000,01.6000,01.8000,02.0000}\draw (\x cm,1.5pt) -- (\x cm, -1.5pt);
            \node (Label) at (00.2000,0.2) {\tiny{1}};
            \node (Label) at (00.4000,0.2) {\tiny{2}};
            \node (Label) at (00.6000,0.2) {\tiny{3}};
            \node (Label) at (00.8000,0.2) {\tiny{4}};
            \node (Label) at (01.0000,0.2) {\tiny{5}};
            \node (Label) at (01.2000,0.2) {\tiny{6}};
            \node (Label) at (01.4000,0.2) {\tiny{7}};
            \node (Label) at (01.6000,0.2) {\tiny{8}};
            \node (Label) at (01.8000,0.2) {\tiny{9}};
            \node (Label) at (02.0000,0.2) {\tiny{10}};
            \draw[decorate,decoration={snake,amplitude=.4mm,segment length=1.5mm,post length=0mm}, very thick, color = black](00.9936,-00.2500) -- ( 01.2403,-00.2500);
            \node (Point) at (01.0436, 0){};  \node (Label) at (0.5,-00.4500){\scriptsize{RF}}; \draw (Point) |- (Label);
            \node (Point) at (01.0568, 0){};  \node (Label) at (0.5,-00.7500){\scriptsize{NB}}; \draw (Point) |- (Label);
            \node (Point) at (01.0636, 0){};  \node (Label) at (0.5,-01.0500){\scriptsize{LR}}; \draw (Point) |- (Label);
            \node (Point) at (01.0703, 0){};  \node (Label) at (0.5,-01.3500){\scriptsize{KNN}}; \draw (Point) |- (Label);
            \node (Point) at (01.0703, 0){};  \node (Label) at (0.5,-01.6500){\scriptsize{CART}}; \draw (Point) |- (Label);
            \node (Point) at (01.1903, 0){};  \node (Label) at (2.5,-00.4500){\scriptsize{Defaults}}; \draw (Point) |- (Label);
            \node (Point) at (01.1703, 0){};  \node (Label) at (2.5,-00.7500){\scriptsize{GP}}; \draw (Point) |- (Label);
            \node (Point) at (01.1606, 0){};  \node (Label) at (2.5,-01.0500){\scriptsize{Random}}; \draw (Point) |- (Label);
            \node (Point) at (01.1036, 0){};  \node (Label) at (2.5,-01.3500){\scriptsize{SVM}}; \draw (Point) |- (Label);
            \node (Point) at (01.0703, 0){};  \node (Label) at (2.5,-01.6500){\scriptsize{Tuning}}; \draw (Point) |- (Label);
        
        \end{tikzpicture}
       \label{fig:friedman_test_j48}
    }
     \caption{Performance of the Meta-learners projected in the J48 hyperparameter tuning problem (base-level).}
    \label{fig:dt_projected_results}
\end{figure}
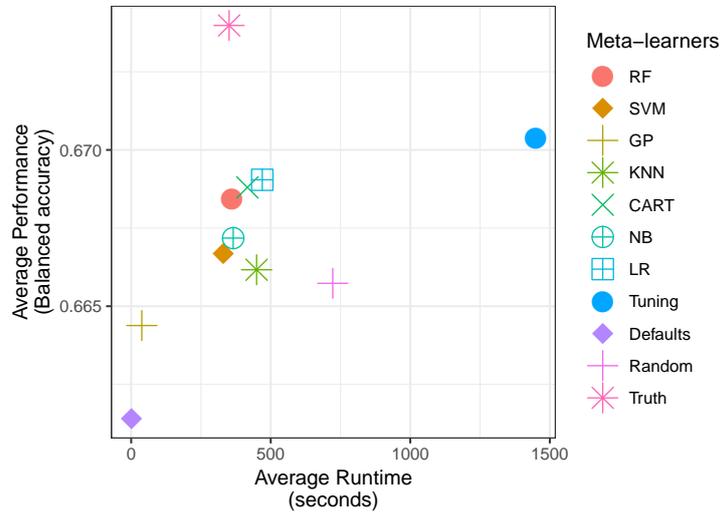
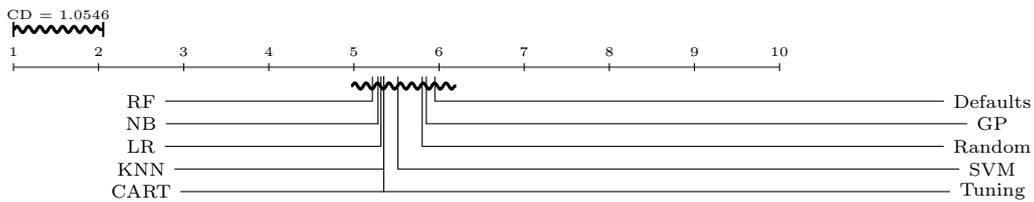




Figure~\ref{fig:dt_projected_results} shows J48 meta-level predictions projected overall base-level datasets. Mostly, but not all of the meta-examples are labeled with ``\texttt{Defaults}''. The induced meta-models depicted in Figure~\ref{fig:j48_projections} are above the ``\texttt{Defaults}'' baseline, but relatively close to the \texttt{Random} and \texttt{Tuning} ones. The average \acrshort{bac} values of all the approaches are very close, even considering all the baselines. This can be noted by the scale on the y-axis. It is explained by the small improvements obtained at the base-level tuning processes; they were relatively small if compared with those obtained with~\acrshortpl{svm}. 

However, all the meta-models have a lower average runtime compared with \texttt{Random} and \texttt{Tuning} baselines. Most of the meta-models are better ranked than baselines, but, overall, there are no statistical differences among the evaluated approaches (Figure~\ref{fig:friedman_test_j48}). Even so, it is important to highlight that meta-learners could also considerably reduce the computational cost related to tuning, keeping the predictive performance in the dataset collection.


\section{Conclusions} 
\label{sec:conclusions}

This paper proposed and experimentally investigated a \acrshort{mtl} framework to predict when to perform \acrshort{svm} \acrshort{hp} tuning.
To do this, 156 different datasets publicly available at \acrshort{openml} were used. 
The predictive performances of \acrshort{svm} induced with \acrshort{hp} tuning (by a simple \acrshort{rs}) and with default \acrshort{hp} values were compared and used to design a recommender system. 
The default values were provided by the \texttt{e1071} R package), and by optimized common settings from~\cite{Mantovani:2015c}. 
Different experimental setups were analyzed with different sets of meta-features. The main findings are summarized next.


\subsection{Tuning prediction}

The main issue investigated in this paper was whether it is possible to accurately predict when \acrshort{hp} tuning can improve the predictive performance of \acrshortpl{svm}, when compared to the use of default \acrshort{hp} values. 
If so, this can reduce the processing cost when applying \acrshortpl{svm} to a new dataset.
According to experimental results, using \acrshort{rf} and \acrshort{svm} as meta-learners, 
this prediction is possible with a predictive accuracy of \acrshort{auc}=$0.798$.

Three significance levels ($\alpha=\{0.01, 0.05, 0.10\}$) were used with the Wilcoxon test to define the meta-target, 
which indicates whether it is better to use \acrshort{hp} tuning or default \acrshort{hp} values.
Different sets of meta-features were evaluated, and \acrshort{rf} meta-models using all the available meta-features obtained the best results, regardless the $\alpha$ value considered.
However, the complex set of meta-features resulted in high predictive performance for most of the investigated meta-learners.
Different experimental setups were also evaluated at the meta-level, but improvements were observed, and in a few cases, only when \acrshort{smote} or meta-feature selection were used.
Thus, the best setup was to use raw meta-data and meta-learners with default \acrshort{hp} values. 

An analysis of the \acrshort{rf} meta-models show that most meta-features actively contributed to predictions. 
This explains the decrease in performance when using meta-feature selection in most of the algorithms. 
Among the $10$ meta-features ranked as most important, there are meta-features from different sets. 
Each ranked meta-feature describes different characteristics of the problem, such as data imbalance, linearity, and complexity. 

This paper also investigated the hypothesis that the linear separability level could be an important meta-feature to be used by the recommender system. 
Meta-features based on relative landmarking were used to measure the linear separability degree. 
In the experiments performed, this meta-feature was shown to play an important role in the recommender system prediction.
Three meta-learners had their best \acrshort{auc} performance values using a combination of simple and relative landmarking meta-features.

Using two different default \acrshort{hp} settings maximized the number of default wins, reducing the imbalance rate at the meta-datasets. 
In addition to presenting the best predictive performance, \acrshortpl{rf}, by the frequency of meta-features in their trees, provided useful information regarding when default settings are suitable. 

We also performed experiments for the J48 tuning recommendation problem aiming to show the ability of generalization of the \acrshort{mtl} recommender system. Results showed that different to~\acrshortpl{svm}, where the linearity was important to recommend the use of default settings, the most important meta-features suggest that if a dataset has representative attributes, default settings are robust to solve it.

In fact, our extensive experiments suggest that the guideline depends on the algorithm used to induce the meta-model. If we use a white box algorithm, such as \acrshort{rf}, we can use the meta-features in the root of the trees (and nodes close to the root) to explain when to tune. The high predictive performance of the \acrshort{rf} algorithm indicates that the induced models were able to find a good hypothesis for situations where tuning is necessary, in both cases (for \acrshortpl{svm} and J48).


\subsection{Linking findings with the literature}

Two of the related studies in the literature~\cite{Mantovani:2015b,Ridd:2014} used meta-models based on decision trees to interpret their predictions.
However, in the current results, \acrshort{cart} trees were among the worst meta-learners considering all the experimental setups analyzed. 
Thus, in this study, the meta-analysis performed was based on the \acrshort{rf} meta-learner, extracting the average of the Gini index from the meta-features provided by the inner \acrshort{rf} meta-models.

Meta-feature selection was also evaluated in~\cite{Ridd:2014}. 
The authors explored a \acrfull{cfs} method and reported the ``\texttt{nn}'' meta-feature\footnote{The performance of 1-NN algorithm. See tables~\ref{tab:mfeats_1} and \ref{tab:mfeats_2} in Appendix~\ref{app:features}.} as the most important. 
However, the results reported here were not improved by meta-feature selection. 
In fact, it decreased the performance of most meta-models, as shown in Figure~\ref{fig:perf}(a).
Meta-feature selection was also tried with filter methods, but the results were even worse than using \acrshort{sfs}. 
For this reason, it was not reported in this study.
In addition, ``\texttt{nn}'' meta-features did not appear among the top-20 most important meta-features computed by \acrshort{rf}.

Our experimental results show that using meta-features from different categories have improved the predictive performance of the meta-learners for different setups.
The most important meta-features were ''\texttt{SM.classes\_min}'' and ''\texttt{LM.stump\_sd}''\footnote{This is shown in Figure~\ref{fig:rf-mfeat} in Subsection~\ref{subsec:rf}.}. 
In ~\cite{Mantovani:2015b}, which only used  meta-features from simple and data complexity sets, ``\texttt{SM.classes\_max}'' and ``\texttt{SM.attributes}'' were reported as the most important meta-features.
The first describes the percentage of examples in the majority class. The second is the number of predictive attributes in a dataset. 
Although these studies disagree about the relative importance order of these meta-features, both extract information related to the same characteristics: data complexity and dimensionality.

The \acrshort{hp} tuning investigated in~\cite{Sanders:2017}
``\textit{assumed that tuning is always necessary}'', and therefore focused on the improvement prediction as a regression task. They obtained \acrshort{hp} tuning results for less than half of the datasets (111/229) in the base-level when dealing with \acrshortpl{svm}. In addition, their meta-models were not able to predict the \acrshort{hp} improvement for \acrshortpl{svm}, not providing any valid conclusion about the problem.


\subsection{Main difficulties}

During the experiments, there were several difficulties to generate the meta-knowledge. 
The process itself is computationally expensive, since a lot of tuning tasks must be run and evaluated in a wide range of classification tasks. Initially, a larger number of datasets were selected, but some of them were extremely expensive computationally speaking for either \acrshort{hp} tuning or the extraction of meta-features. 
A \textit{walltime} of $100$ hours was considered to remove high-cost datasets. 

The class imbalance at the meta-level was another problem faced in the experiments.
To deal with this problem, the optimized default \acrshort{hp} settings~\cite{Mantovani:2015c} were added to the meta-dataset, 
increasing the number of meta-examples in the default meta-target.
Besides the addition of relative landmarking set, some meta-examples were never correctly classified.
This points out the need to define specific meta-features for some \acrshort{mtl} problems.


\subsection{Future work}

Some findings from this study also open up future research directions. 
The proposed MtL recommender system could be extended to different \acrshort{ml} algorithms, such as neural networks, another decision tree induction algorithms and ensemble-based techniques. 
It could also be used to support \acrshort{hp} tuning decision in different tasks, such as pre-processing, regression and clustering. 
It could even be used to define whether to tune \acrshort{hp} for more than one task, in a pipeline or simultaneously.

It would also be a promising direction to investigate the need of new meta-features to characterize data when dealing with data quality problems, for instance imbalancing measures, due to their influence in the quality of the induced meta-models.
Besides, a multicriteria objective function could replace the current meta-label rule, weighing predictive performance, memory, and runtime.
Another possibility would be to explore the use of ensembles as meta-models, given the complementary behavior of some of the algorithms studied here as meta-learners. 

The code used in this study is publicly available, easily extendable and may be adapted to cover several other \acrshort{ml} algorithms. 
The same may be said of the analysis, also available for reproducibility. 
All the experimental results generated are also available at \acrshort{openml} correspondent studies web pages, from which they can be integrated and reused in different \acrshort{mtl} systems. 
This framework is expected to be integrated to \acrshort{openml}, so that the scientific community can use it.


\section*{Acknowledgments}

The authors would like to thank CAPES and CNPq (Brazilian Agencies) for their financial support, specially for grants \#2012/23114-9, \#2015/03986-0 and \#2018/14819-5 from the S{\~a}o Paulo Research Foundation (FAPESP).

\section*{\refname}


\appendix
\newpage

\section{List of Meta-features used in experiments} 
\label{app:features}

\begin{table*}[h!]
\scriptsize
\centering

\caption{Meta-features used in experiments - part 1. For each meta-features it is shown: its type, acronym and description. Extended from~\cite{Garcia:2016}.}
\label{tab:mfeats_1}
\begin{tabular}{lll}
  
  \toprule
  \multicolumn{1}{l}{\textbf{Type}} & \textbf{Acronym} & \textbf{Description} \\
  \midrule
  
  \multirow{10}{*}{\acrfull{sm}} & classes & Number of classes \\
  & attributes & Number of attributes \\
  & numeric & Number of numerical attributes \\
  & nominal & Number of nominal attributes \\
  & samples & Number of examples \\
  & dimension & samples/attributes \\
  & numRate & numeric/attributes \\
  & nomRate & nominal/attributes \\
  & symbols (min, max, mean, sd, sum) & Distributions of categories in attributes \\  
  & classes (min, max, mean, sd) & Classes distributions \\
  \rule{0pt}{6ex} 

  \multirow{7}{*}{\acrfull{st}} & sks & Skewness \\
  & sksP & Skewness for normalized dataset \\
  & kts  & Kurtosis \\
  & ktsP & Kurtosis for normalized datasets \\
  & absC & Correlation between attributes \\
  & canC & Canonical correlation between matrices \\
  & frac & Fraction of canonical correlation \\
  \rule{0pt}{6ex}  

  \multirow{9}{*}{\acrfull{in}} & clEnt   & Class entropy \\
  & nClEnt  & Class entropy for normalized dataset \\
  & atrEnt  & Mean entropy of attributes \\
  & \multirow{2}{*}{nAtrEnt} & Mean entropy of attributes for\\
  & & normalized dataset \\ 
  & jEnt    & Joint entropy \\
  & mutInf  & Mutual information \\
  & eqAtr   & clEnt/mutInf \\
  & noiSig  & (atrEnt - mutInf)/MutInf \\
    \rule{0pt}{6ex}
  
  \multirow{8}{*}{\acrfull{mb} (Trees)} & nodes  & Number of nodes \\
  & leaves   & Number of leaves \\
  & nodeAtr & Number of nodes per attribute \\
  & nodeIns & Number of nodes per instance \\
  & leafCor & leave/samples \\
  & lev (min, max, mean, sd) & Distributions of levels of depth \\
  & bran (min, max, mean, sd) & Distributions of levels of branches \\
  & att (min, max, mean, sd) & Distributions of attributes used \\
    \rule{0pt}{6ex}
  
  \multirow{5}{*}{\acrfull{lm}} & nb & Naive Bayes accuracy \\
  & stump (min, max, mean, sd) & Distribution of decision stumps \\
  & stMinGain & Minimum gain ratio of decision stumps \\
  & stRand & Random gain ratio of decision stumps \\
  & nn & 1-Nearest Neighbor accuracy \\
  
  \bottomrule

\end{tabular}
\end{table*}

\clearpage

\begin{table*}[h!]
\scriptsize
\centering
\caption{Meta-features used in experiments - part 2. For each meta-features it is shown: its type, acronym and description. Extended from~\cite{Garcia:2016}.}
\label{tab:mfeats_2}
\begin{tabular}{lll}
  
  \toprule
  \multicolumn{1}{l}{\textbf{Type}} & \textbf{Acronym} & \textbf{Description} \\
  \midrule

  \multirow{15}{*}{\acrfull{dc}} & f1  & Maximum Fisher's discriminant ratio \\
  & f1v & Directional-vector maximum Fisher's discriminant ratio \\ 
  & f2  & Overlap of the per-class bounding boxes \\ 
  & f3  & Maximum feature efficiency \\ 
  & f4  & Collective feature efficiency \\ 
  & l1  & Minimized sum of the error distance of a linear classifier \\ 
  & l2  & Training error of a linear classifier \\ 
  & l3  & Nonlinearity of a linear classifier \\ 
  & n1  & Fraction of points on the class boundary \\ 
  & n2  & Ratio of average intra/inter-class NN distance \\ 
  & n3  & leave-one-out error rate of the 1-NN classifier \\ 
  & n4  & Nonlinearity of the 1-NN classifier \\ 
  & t1  & Fraction of maximum covering spheres \\ 
  & t2  & Average number of points per dimension \\
    \rule{0pt}{6ex}
    
  \multirow{9}{*}{\acrfull{cn}} & edges   & Number of edges \\ 
  & degree  & Average degree of the network \\ 
  & density & Average density of the network \\ 
  & maxComp & Maximum number of components \\ 
  & closeness & Closeness centrality \\ 
  & betweenness & Betwenness centrality \\ 
  & clsCoef & Clustering Coefficient \\ 
  & hubs & Hub score \\ 
  & avgPath & Average path length \\
    \rule{0pt}{6ex}

  \multirow{10}{*}{\acrfull{rl}} & diff.svm.lm & performance(SVM) - performance(Linear) \\
  
  & diff.svm.nb & performance(SVM) - performance(NB)\\ 
  
  & diff.svm.stump & performance(SVM) - performance(Decision Stump) \\
  
  & diff.svm.nn & performance(SVM) - performance(1-NN)\\ 
  
  & diff.nn.lm & performance(1-NN) - performance(Linear) \\
  
  & diff.nn.stump & performance(1-NN) - performance(Decision Stump)\\
  
  & diff.nn.nb & performance(1-NN) - performance(NB)\\
  
  & diff.nb.stump & performance(NB) - performance(Decision Stump)\\
  
  & diff.nb.lm & performance(NB) - performance(Linear)\\
  
  & diff.stump.lm & performance(Decision Stump) - performance(Linear)\\

  \bottomrule

\end{tabular}
\end{table*}
\clearpage
\section{Hyperparameter space of the meta-learners used in experiments} 
\label{app:mtl_space}

\begin{table*}[th!]
\scriptsize
    \centering
    \caption{Meta-learner's hyperparameter spaces explored in the experiments. The nomenclature follows their respective R packages. The \acrshort{nb} and \acrshort{lr} classifiers do not have any hyperparameter for tuning.}
    \label{tab:app_space}
    \begin{tabular}{cclcccc}
            
            \toprule
            
            \textbf{ Algo } & \textbf{ Symbol } & \textbf{hyperparameter} & \textbf{Range} & \textbf{Type} & \textbf{Default} & \textbf{Package} \\
            
            \midrule
            \rule{0pt}{1ex}
          
            \multirow{8}{*}{CART} & cp & complexity parameter & $(0.0001, 0.1)$ & real & $0.01$ & \multirow{8}{*}{\texttt{rpart}} \\
            \rule{0pt}{4ex}
          
            &  \multirow{2}{*}{minsplit} & minimum number of instances in a &  \multirow{2}{*}{$[1,50]$} & \multirow{2}{*}{integer} & \multirow{2}{*}{$20$} & \\
            & & node for a split to be attempted \\
             \rule{0pt}{4ex}
          
            & minbucket & minimum number of instances in a leaf & $[1,50]$ & integer & $7$ &  \\
           \rule{0pt}{4ex}
            
            & \multirow{2}{*}{maxdepth} & maximum depth of any node of & \multirow{2}{*}{$[1,30]$} & \multirow{2}{*}{integer} & \multirow{2}{*}{$30$} & \\
            & & the final tree \\
            \rule{0pt}{7ex}
            
            GP & sigma & width of the Gaussian kernel & $[2^{-15}, 2^{15}]$ & real & - &  \texttt{kernlab} \\
            
             \rule{0pt}{7ex}

            \multirow{5}{*}{SVM} & k & kernel & Gaussian & - & - & \multirow{5}{*}{\texttt{e1071}} \\ \rule{0pt}{4ex}
            & C & regularized constant & $[2^{-15},2^{15}]$ & real & 1 \\
             \rule{0pt}{4ex}
            &$\gamma$ & width of the Gaussian kernel & $[2^{-15}, 2^{15}]$ & real & $\rfrac{1}{N}$ & \\
            
            \rule{0pt}{7ex}            
           
            \multirow{2}{*}{RF} & ntree & number of trees & $[2^0,2^{10}]$ & integer & 500 & \multirow{3}{*}{\texttt{randomForest}} \\
             \rule{0pt}{4ex}
            
            & nodesize & minimum node size of the decision trees & $\{1,20\}$ & integer & 1 & \\
            
            \rule{0pt}{7ex}
            
            KNN & k & number of nearest neighbors & $\{1,50\}$ & integer & 7 & \texttt{kknn} \\

            \rule{0pt}{5ex}
            
            NB & - & - & - & - & - & \texttt{e1071} \\
            
            \rule{0pt}{5ex}            
            
            LR & - & - & - & - & - & \texttt{gbm} \\

            \bottomrule
            
        \end{tabular}
\end{table*}
\clearpage
\section{J48 hyperparameter space and meta-datasets used in experiments from Section~\ref{sec:generalization}} 
\label{app:j48_space}


\begin{table*}[th!]
\scriptsize
    \centering
    \caption{J48 hyperparameter space explored in experiments. The nomenclature is based on the \texttt{RWeka} package. Table adapted from~\cite{Mantovani:2016}.}
    \label{tab:dt_hyperspace}
    \begin{tabular}{clcccc}
            
            \toprule
            
            \textbf{Symbol} & \textbf{Hyperparameter} & \textbf{Range} & \textbf{Type} & \textbf{Default} & \textbf{Conditions} \\
            
            \midrule
            \rule{0pt}{1ex}
          
            C & pruning confidence & $(0.001, 0.5)$ & real & 0.25 & R = False \\
            \rule{0pt}{4ex}
          
            M & minimum number of instances in a leaf & $[1,50]$ & integer & 2 & - \\
            \rule{0pt}{4ex}
            
            \multirow{2}{*}{N} & number of folds for reduced & \multirow{2}{*}{$[2,10]$} & \multirow{2}{*}{integer} & \multirow{2}{*}{3} & \multirow{2}{*}{R = True} \\
            & error pruning & & & & \\
            \rule{0pt}{4ex}
           
            O & do not collapse the tree & \{False,\:True\} & logical & False & - \\
            \rule{0pt}{4ex}
          
            R & use reduced error pruning & \{False,\:True\}  & logical   & False & - \\
           \rule{0pt}{4ex}
            
            B & use binary splits only & \{False,\:True\} & logical & False & - \\
            \rule{0pt}{4ex}
            
            S & do not perform subtree raising & \{False,\:True\} & logical & False & - \\
            \rule{0pt}{4ex}
           
            \multirow{2}{*}{A} & Laplace smoothing for predicted & \multirow{ 2}{*}{\{False,\:True\}}    & \multirow{2}{*}{logical}   & \multirow{ 2}{*}{False} & - \\
            &  probabilities & & & & \\
           \rule{0pt}{4ex}
            
            \multirow{2}{*}{J} & do not use MDL correction for & \multirow{ 2}{*}{\{False,\:True\}} & \multirow{2}{*}{logical} & \multirow{2}{*}{False} & - \\
            & info gain on numeric attributes & & & & \\
            
            \bottomrule
    \end{tabular}
\end{table*}


\begin{table}[h!]
    \caption{Meta-datasets generated for J48 experiments.}
    \label{tab:j48_datasets}
    \centering
    \begin{tabular}{crcccc}
    
    \toprule
    \multirow{2}{*}{\textbf{Meta-dataset}} & \multirow{2}{*}{\textbf{$\alpha$}}
    & \textbf{Meta} & \textbf{Meta} & \multicolumn{2}{c}{\textbf{Class Distribution}} \\
    & & \textbf{examples} & \textbf{features} & \textit{Tuning} & \textit{Default} \\
    \midrule
    
    \rule{0pt}{1ex} 

    J48\_90 & $0.10$ & 165 & 80 & 63 & 102 \\
    \rule{0pt}{1ex} 
    J48\_95 & $0.05$ & 165 & 80 & 57 & 108 \\
    \rule{0pt}{1ex} 
    J48\_99 & $0.01$ & 165 & 80 & 52 & 113 \\
    
    \bottomrule

    \end{tabular}
\end{table}

\clearpage
\section{List of abbreviations used in the paper} 
\label{app:glossary}

\printnoidxglossary[type=\acronymtype, title=\empty]

\end{document}